\documentclass{article}

\usepackage{PRIMEarxiv}

\usepackage[utf8]{inputenc} 
\usepackage[T1]{fontenc}    
\usepackage{hyperref}       
\usepackage{url}            
\usepackage{booktabs}       
\usepackage{amsfonts}       
\usepackage{nicefrac}       
\usepackage{microtype}      
\usepackage{lipsum}
\usepackage{fancyhdr}       
\usepackage{graphicx}       
\graphicspath{{media/}}     

\usepackage{amsmath,amsfonts}
\usepackage{algorithm}
\usepackage{array}
\usepackage[caption=false,font=normalsize,labelfont=sf,textfont=sf]{subfig}
\usepackage{textcomp}
\usepackage{stfloats}
\usepackage{url}
\usepackage{verbatim}
\usepackage{graphicx}
\usepackage{cite}
\usepackage{amssymb}
\usepackage{subfig}
\usepackage{textcomp}
\usepackage{xcolor}
\usepackage{todonotes}
\usepackage{algpseudocode}
\usepackage{multirow}
\usepackage{pifont}

\usepackage[inline]{enumitem}
\title{Federated Self-Supervised Learning of Monocular Depth Estimators for Autonomous Vehicles
}

\author{Elton F. de S. Soares \\
  IBM Research \& UNIRIO \\
  Rio de Janeiro\\[3pt] 
  {\tt eltons@ibm.com, elton.soares@uniriotec.br}\\
   \And
  Carlos Alberto V. Campos \\
  UNIRIO\\
  Rio de Janeiro\\
  {\tt beto@uniriotec.br}\\
}

\begin{document}
\maketitle

\begin{abstract}
Image-based depth estimation has gained significant attention in recent research on computer vision for autonomous vehicles in intelligent transportation systems. This focus stems from its cost-effectiveness and wide range of potential applications. Unlike binocular depth estimation methods that require two fixed cameras, monocular depth estimation methods only rely on a single camera, making them highly versatile. While state-of-the-art approaches for this task leverage self-supervised learning of deep neural networks in conjunction with tasks like pose estimation and semantic segmentation, none of them have explored the combination of federated learning and self-supervision to train models using unlabeled and private data captured by autonomous vehicles. The utilization of federated learning offers notable benefits, including enhanced privacy protection, reduced network consumption, and improved resilience to connectivity issues. To address this gap, we propose FedSCDepth, a novel method that combines federated learning and deep self-supervision to enable the learning of monocular depth estimators with comparable effectiveness and superior efficiency compared to the current state-of-the-art methods. Our evaluation experiments conducted on Eigen's Split of the KITTI dataset demonstrate that our proposed method achieves near state-of-the-art performance, with a test loss below 0.13 and requiring, on average, only 1.5k training steps and up to 0.415 GB of weight data transfer per autonomous vehicle on each round.
\end{abstract}

\keywords{Monocular Depth Estimation \and Self-Supervised Learning \and Federated Learning}

\section{Introduction}
\label{sec:intro}

Because of the adverse impact of a poorly managed mobility system on the quality of life, Smart Mobility is often presented as one of the main options to seek more sustainable transport systems~\cite{khamis2021smart}. It could also be seen as a set of coordinated actions aimed at improving cities' efficiency, effectiveness, and environmental sustainability.
One of these actions is the development of Intelligent Transportation Systems (ITS), which has been occurring since the beginning of the 1970s and can be seen as the integration of advanced technologies, which include electronic sensor technologies, data transmission technologies, and intelligent control technologies, into the transportation systems~\cite{zhu2018big}. Nonetheless, the primary purpose of ITS is to provide better services for drivers and riders~\cite{an2011survey}.

In the last few years, a large amount of research effort has been made to apply Big Data Analytics and other advanced Artificial Intelligence (AI) techniques to improve ITS~\cite{hu2017intelligent}. In contrast, a smaller amount has been focused on developing intelligent agents to support ITS.
The primary efforts made in that sense are those focused on developing Autonomous Vehicles (AVs), which are now one of the most prominent topics in the ITS initiative~\cite{aldakkhelallah2021autonomous}.
Research on AVs has also applied advanced AI techniques to tackle its most critical tasks, such as Computer Vision (CV).
Scene Depth Estimation (DE) plays an essential role in CV as it enables the perception and understanding of three-dimensional scenes~\cite{ming2021deep}.
Lasers, structured light, and other reflections on the object surface have traditionally been applied in active DE methods~\cite{ming2021deep}.
To enable these approaches, elevated costs of human labor and computational resources are usually required for obtaining dense and accurate depth maps~\cite{zhang2020monocular}.

Thus, image-based DE has become one of the main focuses of recent research in CV for AVs due to its lower deployment cost and a wider range of application scenarios~\cite{ming2021deep}.
Image-based DE methods traditionally calculate the disparity between two 2D images a binocular camera takes to obtain a depth map~\cite{gorban2020deep}.
However, binocular DE methods require at least two fixed cameras, and it is difficult to capture enough features in the image to match when the scene has less or no texture~\cite{liu2019binocular}.

Therefore, research began focusing on Monocular DE (MDE)~\cite{laga2019survey}.
Since MDE uses a single camera to obtain an image or video sequence, which does not require additional specialized equipment, it has an even wider applicability~\cite{ming2021deep}. Nonetheless, as monocular images lack a reliable stereoscopic visual relationship, the regression of depth in 3D space from it is an ill-posed problem~\cite{ming2021deep}.
More specifically, monocular images adopt a 2D form to reflect the 3D world. However, the depth of the scene is not captured by the imaging process, making it impossible to judge the size and distance of an object in the scene or whether it is occluded by another object~\cite{ming2021deep}.

Thus, we need to estimate the depth of each pixel from the monocular image. Based on the pixel depth map, we can judge the size and distance of the objects contained in that scene. When the estimated depth map can accurately reflect the 3D structure of the scene, we can consider the estimation method used to be effective~\cite{ming2021deep}.
Several State-of-The-Art (SoTA) solutions for MDE make use of Self-Supervised Learning (SSL) of Deep Neural Networks (DNNs) for this task in combination with other CV tasks, such as ego-motion/pose estimation (PE) and semantic segmentation (SS)~\cite{jing2020self, liu2022self}.
Nonetheless, to the best of our knowledge, none of the SoTA solutions for MDE combines the use of Federated Learning (FL)~\cite{mcmahan2017communication} with SSL to learn MDE models from unlabeled and private data captured by AVs.

The use of FL has been explored in many recent works on ITS and AVs~\cite{manias2021making, du2020federated}. The main advantages of FL~\cite{savazzi2021opportunities} include: (1) increased privacy protection, as there is no longer the need to share the raw data collected by each vehicle with a central server or other vehicles; (2) reduced network consumption, as the size of the model updates that need to be shared in the FL process is significantly smaller than the raw datasets; (3) increased resiliency to connectivity loss when compared to the centralized approach; and (4) increased robustness to Non-IID (independent and identically distributed) data~\cite{ma2022state}. Thus, we hypothesize that combining FL and SSL can enable learning models with comparable effectiveness and superior efficiency to the SoTA methods in MDE for AVs.
Also, several works have explored the combination of SSL and FL on CV tasks with promising results~\cite{zhang2020federated, zhao2021hotfed, zhuang2021collaborative, li2022fedutn, wang2022does, makhija2022federated, shi2022fedcoco, wu2022decentralized, zhuang2022divergence, qu2022rethinking, mu2023fedproc, park2023ms, yang2023fedil, yang2023fedmae, zhao2023fedusc}.
Nonetheless, none of them were evaluated on datasets of images collected by vehicles, such as the SoTA benchmarks for MDE models~\cite{kittydepth}.

Thus, this work's main objective is to develop a solution for the problem of MDE for AVs. This solution must be able to generate depth maps of images captured by monocular cameras in moving AVs with high effectiveness and efficiency. 
MDE effectiveness is essential for scene understanding by AVs, as the depth information will help identify the distance of obstacles as well as estimate the speed and acceleration of other moving vehicles~\cite{jin2020practical}. Meanwhile, high efficiency is another critical requirement of the ideal solution because it cannot consume a high proportion of the computational resources available on the vehicle, as these are already disputed by the other tasks the vehicle must do in real-time. Besides, the ITS network infrastructure might be unable to support the sharing of all the training data between AVs and/or a central server; therefore, mitigating the bandwidth consumption can increase the ITS infrastructure's scalability.

To the best of our knowledge, this is the first work to present and discuss empirical evidence of the applicability of Self-Supervised Federated Learning (SSFL) to MDE for AVs.

This work tackles the following Research Questions (RQs):
\begin{itemize}
\item {\it RQ1 -} Is the \textit{efficiency} of the SSL of MDE models \textit{higher} when applying FL (with IID and Non-IID data) or a centralized approach~\footnote{Training a model on a central server with data collected by all vehicles.}, in the AVs use case?
\item {\it RQ2 -} Is the \textit{effectiveness} of SSL MDE models \textit{equivalent} when applying FL (with IID and Non-IID data) instead of a centralized approach in the AVs use case?
\end{itemize}

To answer the RQs, we provide the following contributions:

\begin{enumerate}\item We propose the FedSCDepth method to solve the problem of MDE in AVs using SSFL for collaboratively training a depth estimator using unlabeled data captured by vehicles with high effectiveness, efficiency, and privacy;
\item We present an empirical evaluation of a prototype of the proposed method using a real dataset for MDE in AVs.
\item We show that FedSCDepth reaches comparable performance with the SoTA on MDE, with lower computation and communication costs per vehicle per round than centralized training, using both IID and Non-IID data;
\end{enumerate}

Section~\ref{sec:relatedwork} discusses the related work. Section~\ref{sec:proposal} presents the proposed method, and Section~\ref{sec:experiments} details the evaluation experiments. Section~\ref{sec:discussion} discusses the results obtained, and Section~\ref{sec:conclusion} presents conclusions and future work.

\section{Related Work}
\label{sec:relatedwork}

In this section, we present the theoretical background of the methods proposed for solving the MDE problem, the methods that leveraged FL in the ITS and AV domains, and the recent works that combined SSL and FL for CV tasks.

\subsection{Evolution of Monocular Depth Estimation Methods}

During the early phase of DE research, depth maps were primarily estimated using various depth cues such as vanishing points, focus and defocus, and shadows. However, most of these methods were limited to constrained scenes~\cite{ming2021deep}.
In the subsequent Machine Learning (ML) period of DE research, researchers proposed several handcrafted features and probabilistic graph models.
These models were utilized for MDE using parametric and non-parametric learning within the ML framework~\cite{ming2021deep}.
The emergence of Deep Learning (DL) marked a new period in DE research in which MDE became a task of inferring depth maps from single 2D color images using DNNs. Eigen et al.~\cite{eigen2014depth} pioneered this approach by introducing a coarse-to-fine framework.

DL techniques for MDE commonly employ encoder-decoder networks to generate depth maps from RGB images. The encoder captures depth features using convolution and pooling layers, while the decoder estimates pixel-level depth maps using deconvolution layers. Skip connections preserve features at different scales. Training involves minimizing a depth loss function until a predefined threshold is reached~\cite{ming2021deep}. Gradient descent variants are commonly used, but their quality depends on hyperparameters and network initialization. Image resizing is often necessary during initialization.

Supervised and semi-supervised approaches to MDE will typically require some amount of labeled data~\cite{ming2021deep}, which might not represent training truly general models for DE in the heterogeneous domains where AVs will be deployed.
To address this problem, several unsupervised methods have been proposed for learning visual features from large datasets of unlabeled images or videos without relying on human annotations~\cite{jing2020self}. These methods, often called self-supervised, utilize pseudo-labels generated from raw data. Typically, they employ one or more pretext tasks to learn from unlabeled data. By optimizing the objective functions of pretext tasks, DNNs acquire higher-order representational features, enabling them to predict desired visual features such as image depth~\cite{jing2020self}.

\subsection{Self-Supervised Learning for Monocular Depth Estimation}

SSL has introduced various pretext tasks, including colorizing grayscale images, image inpainting, and image jigsaw puzzles~\cite{ming2021deep}. These pretext tasks have been explored in conjunction with other training paradigms. Similarly, in addition to single-task learning, which involves training a single network for DE, combining DE with other tasks such as PE, SS, and optical flow prediction can lead to the acquisition of shared representations beneficial for multiple related tasks~\cite{ming2021deep, jing2020self}.

A notable series of works that incrementally enhanced SSL for MDE was the SC-Depth series methods~\cite{bian2019neurips, bian2021tpami, sc_depthv3}. In SC-DepthV1~\cite{bian2019neurips}, authors focused on the scale inconsistency issue of preexisting solutions and proposed a method to enable scale-consistent DE over video. In their following work, SC-DepthV2~\cite{bian2021tpami}, they focused on the rotation issue in videos that are captured by handheld cameras and proposed an auto-rectify network to handle large rotations. Finally, in SC-DepthV3~\cite{sc_depthv3}, they focused on the issue of dynamic objects and blurred object boundaries. Provided that, they proposed a method that leverages an externally pretrained MDE model for generating single-image depth prior, namely pseudo-depth, based on which novel losses are computed to boost SSL. As a result, the models trained through this method can predict sharp and accurate depth maps, even when trained from monocular videos of highly dynamic scenes.

In the present work, we use SC-DepthV3~\cite{sc_depthv3} as our baseline method for centralized SSL of MDE models since it presented great results on two popular benchmarking datasets for MDE in AVs: KITTI~\cite{geiger2013vision} and DDAD~\cite{guizilini20203d}. Additionally, the well-documented source code\footnote{https://github.com/JiawangBian/sc\_depth\_pl} provided by its authors enabled us to quickly reproduce their experiments and integrate them within our FL solution.Besides SC-DepthV3, we will also compare our results with DepthFormer~\cite{guizilini2022multi} and MonoFormer~\cite{bae2023deep}, two recent transformer-based method that, to the best of our knowledge, currently hold the best results on KITTI Eigen's Split among the SSL-based methods. DephFormer's main characteristic is that it performs multi-frame SSL-based MDE by improving feature matching across images during cost volume generation~\cite{guizilini2022multi}, while MonoFormer uses a CNN-Transformer hybrid network to increase shape bias by employing Transformers while compensating for the weak locality bias of Transformers by adaptively fusing multi-level representations~\cite{bae2023deep}.

\subsection{Federated Learning (FL)}

Big Data-based ML systems usually collect, clean, and aggregate data into one or multiple central servers deployed in the cloud for model training~\cite{du2020federated}. However, privacy has become a critical aspect of deploying these platforms in recent years. The data used for training typically belongs to different parties that might require different policies and privacy restrictions for sharing data with the platform. In addition, while cloud servers provide highly scalable computational power and storage, transferring data from distributed agents to the cloud might demand high bandwidth from the network infrastructure and incur high communication delays~\cite{du2020federated}.

To tackle these issues, Google proposed FL to allow joint model training by multiple parties~\cite{mcmahan2017communication}. In their approach, the model is assumed to be a neural network whose parameter updates can be shared with a central server without transferring the raw data through the network~\cite{du2020federated}. Usually, the central server, also called Aggregator Agent (AA)~\cite{manias2021making}, orchestrates the training process and determines how often and how many distributed agents, also called Federated Nodes (FN)~\cite{manias2021making}, will contribute to the global model update.

\subsection{Federated Learning with Non-IID data}

The problem of Non-IID data (or heterogeneous data) exists in many ML applications and distributed learning methods~\cite{ma2022state}. ML models are usually trained under the assumption that the training data is IID~\cite{ma2022state}. Thus, when the data of the FL clients or participants is Non-IID with regards to feature values, categorical labels, or even just the quantity of samples, the trained models' performance might be reduced~\cite{ma2022state}.

To comprehend the challenge posed by Non-IID data to FL, we need to consider the SGD algorithm. Many DNN training algorithms depend largely on SGD for optimization~\cite{ma2022state}. SGD updates the gradient of each sample every time~\cite{ma2022state}. Thus, the SGD algorithm converges faster to a local minimum, has a faster update speed, and can be seamlessly applied to FL~\cite{ma2022state}.

In Google’s seminal work~\cite{mcmahan2017communication}, its authors claimed that FedAvg could make FL more robust to Non-IID data, which was put in check by subsequent research that presented evidence that, in some Non-IID data scenarios, FedAvg might be unstable or even divergent~\cite{ma2022state}. Nonetheless, FedAvg is still regarded as a baseline aggregation algorithm for FL, with good results on recent Non-IID data experiments~\cite{wang2022does}.

\subsection{Self-Supervised Federated Learning}

\newcommand{\cmark}{\ding{51}}
\newcommand{\xmark}{\ding{55}}
\def\arraystretch{1.25}
\setlength\tabcolsep{1.5pt}
\begin{table}[htb]

\centering
\caption{Characterization of SSFL works with regards to their applicability to AVs, evaluation on CV tasks, experimentation with IID and NIID, and Datasets used.}
\begin{tabular}{|c|c|c|c|c|l|}
\hline
 \textbf{\begin{tabular}[c]{@{}l@{}}Ref.\end{tabular}} & \textbf{\begin{tabular}[c]{@{}l@{}}AV\end{tabular}} & \textbf{\begin{tabular}[c]{@{}l@{}}CV\end{tabular}} & \textbf{\begin{tabular}[c]{@{}l@{}}IID\end{tabular}} & \textbf{\begin{tabular}[c]{@{}l@{}}NIID\end{tabular}} & \textbf{Datasets} \\ \hline
\cite{saeed2020federated} & \xmark & \xmark & \cmark & \xmark & \begin{tabular}[c]{@{}l@{}}Sleep-EDF, HHAR, MobiAct, WiFi-CSI, WESAD\end{tabular} \\ \hline
\cite{van2020towards} & \xmark & \xmark & \cmark & \xmark & \begin{tabular}[c]{@{}l@{}}HHAR, MobiAct, HAPT\end{tabular} \\ \hline
\cite{servetnyk2020unsupervised} & \xmark & \xmark & \xmark & \cmark & Custom \\ \hline
\cite{zhang2020federated, zhuang2021collaborative} & \xmark & \cmark & \cmark & \cmark & \begin{tabular}[c]{@{}l@{}}CIFAR, Mini-ImageNet\end{tabular} \\ \hline
\cite{zhao2021hotfed, li2022fedutn, zhuang2022divergence} & \xmark & \cmark & \cmark & \cmark & CIFAR \\ \hline
\cite{wang2022does} & \xmark & \cmark & \cmark & \cmark & \begin{tabular}[c]{@{}l@{}}ImageNet, CIFAR, MS-COCO, Amazon\end{tabular}  \\ \hline
\cite{makhija2022federated} & \xmark & \cmark & \cmark & \cmark & \begin{tabular}[c]{@{}l@{}}CIFAR, Tiny-ImageNet, LEAF\end{tabular}  \\ \hline
\cite{shi2022fedcoco} & \xmark & \cmark & \xmark & \cmark & \begin{tabular}[c]{@{}l@{}}CIFAR, SVHN\end{tabular}  \\ \hline
\cite{wu2022decentralized} & \xmark & \cmark & \cmark & \cmark & \begin{tabular}[c]{@{}l@{}}CIFAR, Fashion-MNIST\end{tabular} \\ \hline
\cite{qu2022rethinking} & \xmark & \cmark & \cmark & \cmark & \begin{tabular}[c]{@{}l@{}}Retina, CIFAR, CelebA\end{tabular} \\ \hline
\cite{mu2023fedproc} & \xmark & \cmark & \xmark & \cmark & \begin{tabular}[c]{@{}l@{}}CIFAR, Tiny-ImageNet\end{tabular} \\ \hline
\cite{park2023ms} & \xmark & \cmark & \xmark & \cmark & \begin{tabular}[c]{@{}l@{}}TCIA PET-CT, MICCAI 2015, CTI ICH D\&S\end{tabular} \\ \hline
\cite{yang2023fedil} & \xmark & \cmark & \cmark & \cmark & \begin{tabular}[c]{@{}l@{}}MNIST, CIFAR\end{tabular}  \\ \hline
\cite{yang2023fedmae} & \xmark & \cmark & \cmark & \cmark & \begin{tabular}[c]{@{}l@{}}(Mini-)ImageNet, CIFAR, Mini-INAT2021\end{tabular} \\ \hline
\cite{zhao2023fedusc} & \xmark & \cmark & \cmark & \cmark & \begin{tabular}[c]{@{}l@{}}CIFAR, SVHN, STL-10, COVID-19, Mini-ImageNet \end{tabular} \\ \hline
\textbf{Ours} & \cmark & \cmark & \cmark & \cmark & \begin{tabular}[c]{@{}l@{}}KITTI \end{tabular} \\ \hline
\end{tabular}
\label{tbl:relatedworks}
\end{table}

Several recent works have explored the combination of SSL and FL with promising results. In Table~\ref{tbl:relatedworks}, we characterize those works concerning key aspects, including their applicability for AV use cases, based on the datasets in which they were evaluated.
Although most of them tackled CV tasks~\cite{zhang2020federated, zhao2021hotfed, zhuang2021collaborative, li2022fedutn, wang2022does, makhija2022federated, shi2022fedcoco, wu2022decentralized, zhuang2022divergence, qu2022rethinking, mu2023fedproc, park2023ms, yang2023fedil, yang2023fedmae, zhao2023fedusc}, none of them were evaluated on vehicular datasets, such as the SoTA MDE benchmarks.
Thus, although the approaches proposed by those works could be adapted to AV use cases, none of them provided empirical evidence that SSFL could be adopted successfully in AV use cases, which is precisely the gap we intend to fill in the literature.

Regarding the CV tasks for which SSFL was used, most works focused on image classification tasks. The most frequent datasets were variations of CIFAR~\cite{cifar} and ImageNet~\cite{deng2009imagenet}. Also, most works were evaluated with IID and Non-IID data. Non-IID data is usually generated synthetically based on the number of images containing a given object class. That is another aspect in which the present work differentiates itself since we preserve the natural unbalance of samples inherent to the data collection instead of synthetically generating one based on some assumed distribution skew~\cite{wang2022does}.

\section{Proposed Method}
\label{sec:proposal}

In this section, we detail the proposed method, namely FedSCDepth, which combines an SSL-based MDE component and an FL component, presented in the following subsections.

\subsection{Self-Supervised Monocular Depth Estimation}

In this section, we describe the MDE model and the formalization of frame warping and self-supervision losses.

\subsubsection{Model Architecture}

As in~\cite{sc_depthv3, bian2021tpami, bian2019neurips, godard2017unsupervised}, the core of the model architecture is composed of an MDE network (DepthNet) and a PE network (PoseNet).Both the DepthNet and PoseNet used ResNet18~\cite{laina2016deeper} as their backbone.A fully convolutional U-Net architecture is used for DepthNet~\cite{ronneberger2015u} with a DispNet~\cite{zhou2017unsupervised} as the decoder. The activations are ELU nonlinearities at every layer except the output, where sigmoids are used. The sigmoid output $\sigma$ is converted to depth with $D = \frac{1}{(a\sigma + b)}$, where $a$ and $b$ are chosen to constrain $D$ between $0.1$ and $100$ units~\cite{godard2017unsupervised}. The PoseNet is a ResNet18~\cite{he2016deep} modified to accept a pair of color images (or six channels) as input and to predict a single 6-DoF (Degrees of Freedom) relative pose~\cite{godard2019digging, sc_depthv3}. Also, as proposed in SC-DepthV3~\cite{sc_depthv3} we leverage a pre-trained MDE network (PseudoDepthNet) to generate pseudo-depth. During the training of the DepthNet and PoseNet, the PseudoDepthNet generates a single-image depth prior, which is used to boost SSL.

\begin{figure}[htb]
	\centering
	\includegraphics[trim={0.5cm 0.5cm 0.5cm 0.5cm}, clip, width=0.5\columnwidth]{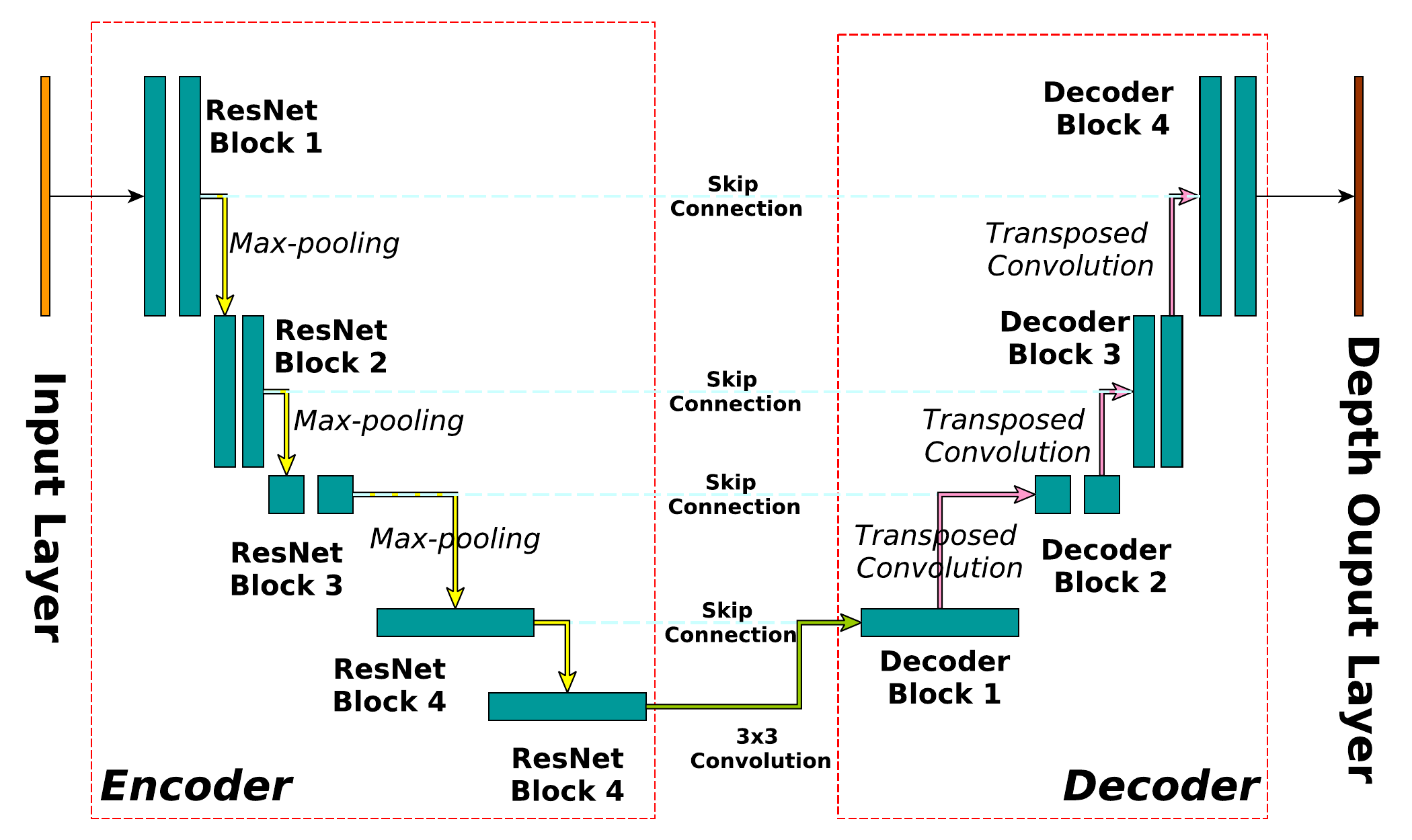}
	\caption{DepthNet architecture. Illustration adapted from~\cite{ronneberger2015u} and~\cite{ahuja2021deep}.}
	\label{fig:depthestimator}
\end{figure}

\begin{figure}[htb]
	\centering
	\includegraphics[trim={0.5cm 0.5cm 0.5cm 0.5cm}, clip, width=0.5\columnwidth]{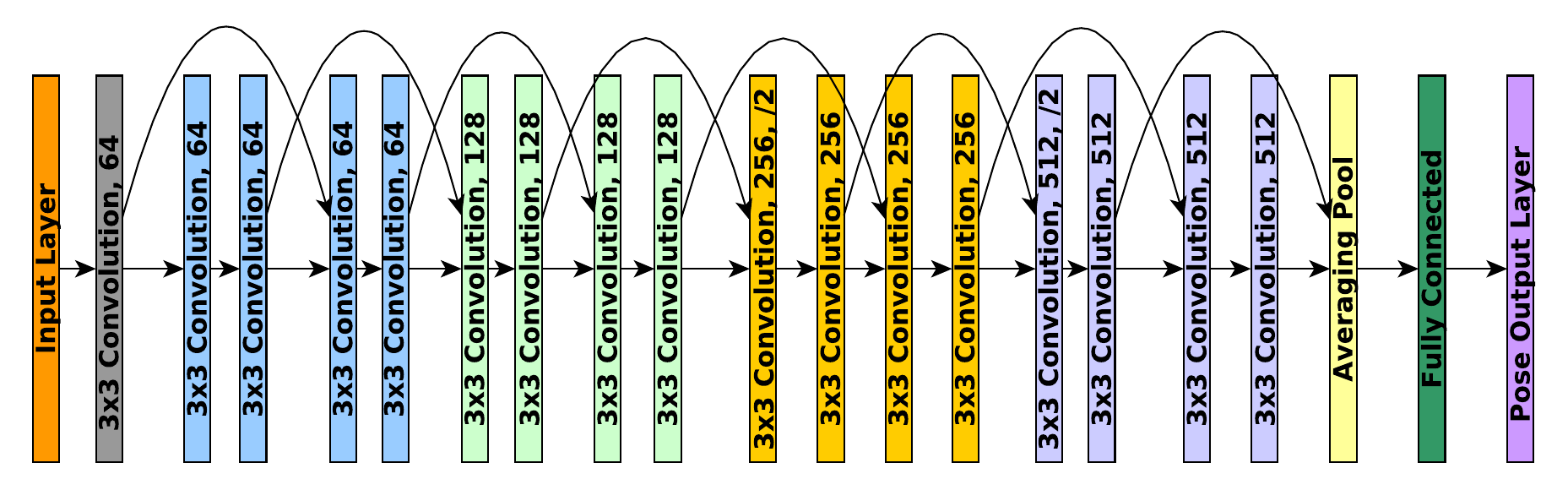}
	\caption{PoseNet architecture. Illustration adapted from~\cite{ramzan2020deep} and~\cite{berga2020disentanglement}. }
	\label{fig:poseestimator}
\end{figure}

\begin{figure}[htb]
	\centering
	\includegraphics[trim={0.5cm 0.5cm 0.5cm 0.5cm}, clip, width=\columnwidth]{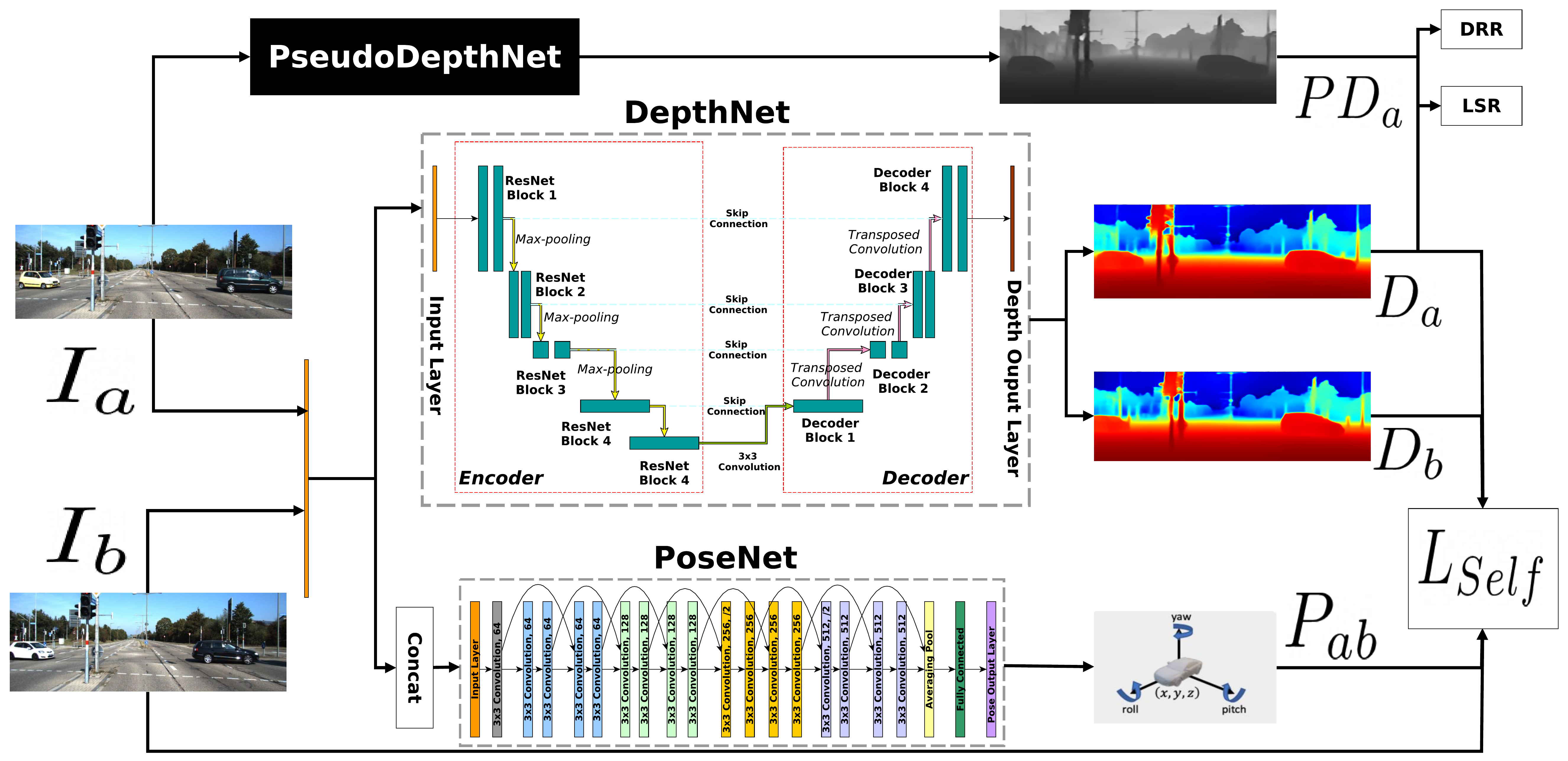}
	\caption{SSL component overview. Given a training sample (\textit{i.e.}, image pair $I_a$ and $I_b$) the combined self-supervision loss ($L_{self}$) is computed. Meanwhile, a pseudo-depth map ($PD_a$) is generated using PseudoDepthNet, while depth maps ($D_a$ and $D_b$) are produced by DepthNet, and PoseNet outputs the pose estimate ($P_ab$). $PD_a$ and $D_a$ are also fed to the Dynamic Region-Refinement (DRR) and Local Structure Refinement (LSR) modules. 
 }
	\label{fig:selfsupervisedlearning}
\end{figure}

An overview of the DepthNet and PoseNet architectures and their combination with PseudoDepthNet in the SSL component is presented in Fig.~\ref{fig:depthestimator}, Fig.~\ref{fig:poseestimator}, and Fig.~\ref{fig:selfsupervisedlearning}, respectively.

\subsubsection{Frame Warping}

Given a sequence of image frames captured by a moving monocular camera, the reconstruction $I'_{s \rightarrow t}$ of a target frame $I_t$ at time $t$ can be obtained from a source frame $I_s$ at time $s$ by performing a bi-linear interpolation over the reprojected frame coordinates. This interpolation, also referred to as warping flow ($W$), can be formalized as~\cite{kuznietsov2021comoda}:
\begin{equation}
    W = I'_{s \rightarrow t}(p_t) = I_s(\hat{p}_s)
\end{equation}
where $\hat{p}_s$ is the reprojection of point $p_t$ into frame $I_s$. To obtain the mapping from $p_t$ to $\hat{p}_s$, $p_t$ needs to be back-projected into 3D point $X$ using the camera's intrinsic matrix $K$ and the depth map $D_t$ corresponding to $I_t$. Then $X$ is transformed to account for camera movement $C_{t \rightarrow s}$ and projected onto the image plain~\cite{kuznietsov2021comoda}. This transformation is formalized as:
\begin{equation}
    \hat{p}_s \sim KC_{t \rightarrow s}\underbrace{D_t(p_t)K^{-1}}_\text{X}.
\end{equation}

\subsubsection{Photometric Loss}

For a consecutive pair of images ($I_a$, $I_b$) randomly sampled from a sequence of monocular images, their depths ($D_a$, $D_b$) and their 6-DoF camera pose ($P_{ab}$) are predicted by forwarding the DepthNet and PoseNet, respectively~\cite{sc_depthv3}.
Provided that, the warping flow ($W_{ab}$) between $I_a$ and $I_b$ can be generated using $D_a$, $D_b$, and $P_ab$, and a synthetic reconstruction of $I_a$ ($I'_a$) can be generated using $W_{ab}$ and $I_b$ via bi-linear interpolation~\cite{zhou2017unsupervised}.
Thus, the photometric loss ($L_p$) between $I_a$ and $I'_a$ can be used as a self-supervision signal for both networks~\cite{bian2019neurips}. Formally,
\begin{equation}
    L_p = \frac{1}{|V|}\sum_{p \in V} ||I_a(p) - I'_a(p)||_1,
\end{equation}
where $V$ corresponds to the valid points that are successfully projected from $I_a$ to the image plane of $I_b$, and $|V|$ stands for the number of points in $V$~\cite{bian2019neurips}. $L_1$ loss is used to reduce the impact of outliers, nonetheless, as it is not invariant to illumination changes, an additional image dissimilarity loss (SSIM~\cite{wang2004image}) is used, as it normalizes the pixel illumination~\cite{bian2019neurips}. The modified $L_p$ is formally defined as,
\begin{equation}
    L_p = \frac{1}{|V|}\sum_{p \in V} (\lambda_i||I_a(p) - I'_a(p)||_1 + \lambda_s \frac{1 - SSIM_{aa'}(p)}{2}),
\end{equation}
where $SSIM_{aa'}$ is the element-wise similarity between $I_a$ and $I'_a$ by the SSIM function~\cite{wang2004image}. $\lambda_i = 0.15, \lambda_s = 0.85$~\cite{godard2017unsupervised}.

\subsubsection{Mask-Weighted Photometric Loss}

To mitigate the adverse impact of moving objects and occlusions, a weight mask $M = 1 - D_{diff}$ is used to assign low weights to inconsistent pixels and high weights to consistent pixels~\cite{bian2019neurips}. Thus, a mask-weighted photometric loss ($L^M_p$) can be formalized as,
\begin{equation}
 L^M_p = \frac{1}{|V|} \sum_{p \in V} (M(p) \cdot L_p(p)).
\end{equation}
By replacing $L_p$ with $L^M_p$, the gradients of inaccurately predicted regions have a lower impact on back-propagation~\cite{bian2019neurips}.

\subsubsection{Combined Self-Supervision Loss Function}

As in~\cite{sc_depthv3}, the edge-aware smoothness loss ($L_s$)~\cite{ranjan2019competitive} is used to regularize the estimated depth maps since $L_p$ is neither very informative in low-texture images nor in homogeneous regions. Also, to enforce that $D_a$ and $D_b$ conform to the same 3D scene structure, another loss was introduced in~\cite{bian2019neurips}, based on a depth inconsistency map. In addition, to mitigate the impact of moving objects and occlusions, a weight mask is used to assign low weights to inconsistent pixels and high weights to consistent ones~\cite{bian2019neurips}. Thus, by replacing $L_p$ with a mask-weighted photometric loss ($L^M_p$), the gradients of inaccurately predicted regions have less impact in back-propagation~\cite{bian2019neurips}.

Finally, as in~\cite{sc_depthv3} the signals produced by the PseudoDetphNet are used to compute additional losses that help regularize the SSL: the Confident Depth Ranking Loss ($L_{cdr}$) and the normal matching loss ($L_n$) that replaces $L_s$~\cite{sc_depthv3}.
Also, the edge-aware relative normal loss ($L_{ern}$) helps constrain the relative normal angles of sampled point pairs to be consistent with pseudo-depth~\cite{sc_depthv3}. Thus, by combining these losses, a robust self-supervision signal is obtained. Formally,
\begin{equation}
L_{Self} = \alpha L^M_p + \beta L_g + \gamma L_n + \delta L_{cdr} + \epsilon L_{ern} .
\end{equation}
As in~\cite{sc_depthv3}, $\alpha = 1$, $\beta = 0.5$, $\gamma = 0.1$. Auto-masking and per-pixel minimum reprojection loss are used to filter stationary and non-best points during training~\cite{godard2019digging} and $\gamma = \delta = \epsilon$~\cite{sc_depthv3}.

\subsection{Federated Learning}

\begin{figure}[H]
	\centering
	\includegraphics[trim={19cm 0 0 0}, clip, width=0.75\columnwidth]{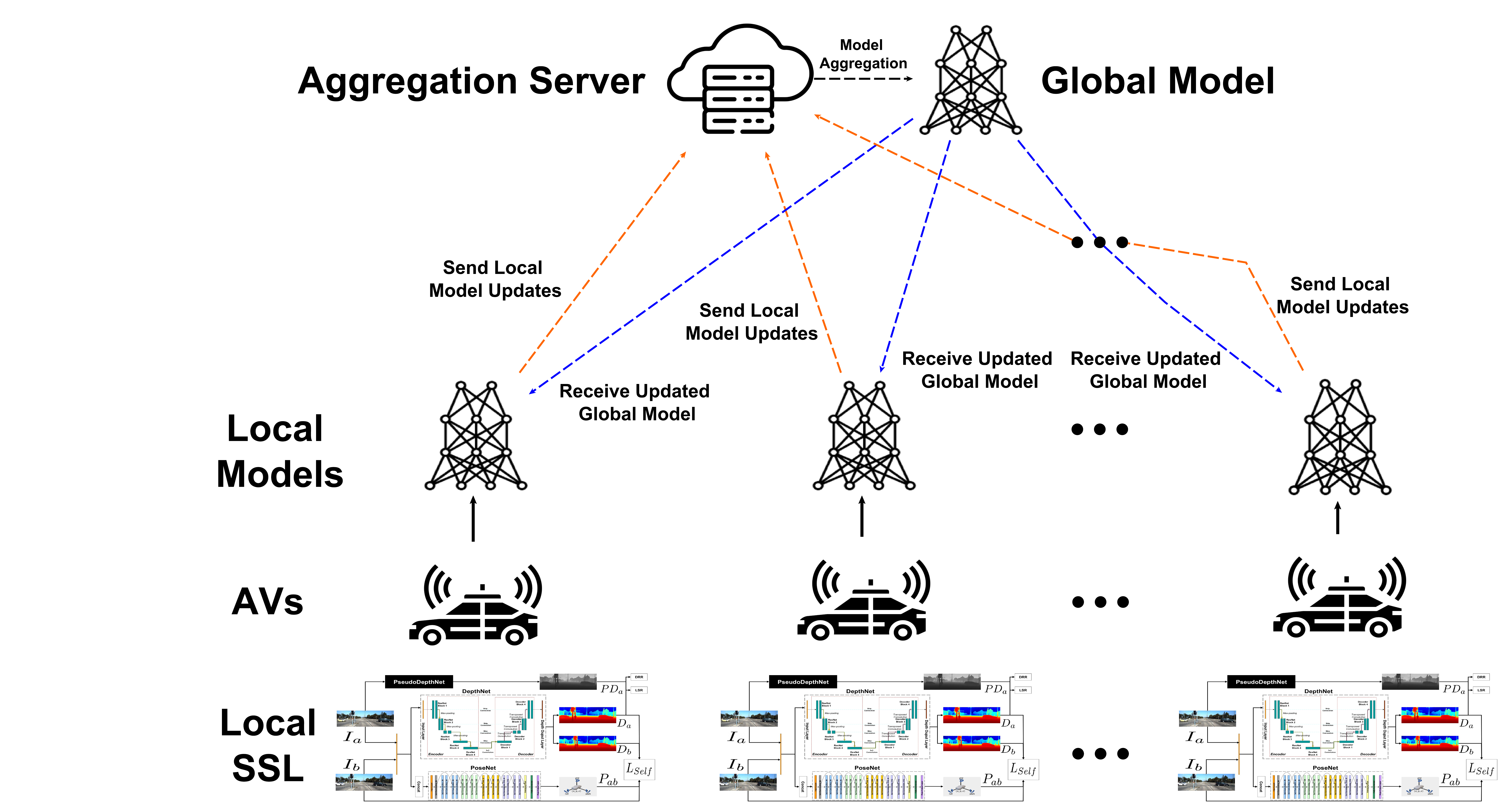}
	\caption{Federated SSL of an MDE estimator for connected AVs. The aggregation server dispatches the initial global models to the participating AVs, as depicted by dashed blue lines. The AVs perform the SSL on their local copies of the global models on their private data and send the model updates back to the server, illustrated with dashed orange lines. The models are aggregated to produce a new version of the global model that is sent back to the AVs. This process is repeated until a stop condition is reached.}
	\label{fig:federatedlearning}
\end{figure}

The main goal of FL is to learn a global model from highly distributed and heterogeneous data by aggregating locally trained models on remote devices~\cite{saeed2020federated}, such as AVs. Considering that our MDE model (DepthNet) is represented as $\varepsilon^\theta_D(I) = D$, our FL goal can be formally defined as:
\begin{equation}
\underset{\theta}{min} \varepsilon^\theta_D, \textnormal{ where } \varepsilon^\theta_D := \sum\limits_{c}^C{\frac{m_c}{m}\varepsilon^\theta_{D_c}} .
\end{equation}
where $C$ represents the number of participating client devices (participants) in an FL round, $m_c$ is the total number of instances available for client $c$ with $m = \sum\nolimits_{c}{m_c}$. Lastly, $\varepsilon^\theta_{D_c}$ denotes the local MDE model parameterized with weights $\theta$. To produce a global model, FedAvg~\cite{mcmahan2017communication} is applied to accumulate client updates after every FL round. Fig.~\ref{fig:federatedlearning} illustrates the FL of a self-supervised MDE model proposed in this work. At the same time, Algorithm~\ref{alg:fedavg} details how FedAvg is applied for generating the global models for MDE and PE.

\begin{algorithm}[H]
\caption{FedAvg of DepthNet and PoseNet. The $C$ participating AVs are indexed by $c$. $F$ is the fraction of AVs active on each FL round. $E$ is the number of training passes each AV makes over its local dataset on each round (local epochs), and $l$ is the number of AVs selected for each round.}\label{alg:fedavg}
\begin{algorithmic}[1]
\setlength{\baselineskip}{0.5\baselineskip}
\item[]
\Require DepthNet $\varepsilon^\theta_D(I) = D$; 
\Require PoseNet $\varepsilon^\theta_P(I_i,I_j) = P_{i \rightarrow j}$;
\item[]
\For{each round $r = 1,2,...$ }
    \item[]
    \State $l = max(F * C, 1)$;
    \State $C_l$ = Random Set of $l$ AVs;
    \item[]
    \For{each AV $c \in C_l$ \textbf{in parallel}}
        \item[]
        \State $\varepsilon^{\theta^c}_{D_{r+1}}$ and $\varepsilon^{\theta^c}_{P_{r+1}} = LocalUpdate(c, \varepsilon^\theta_{D_r},\varepsilon^\theta_{P_r})$;
        \item[]
    \EndFor
    \item[]
    \State $\varepsilon^{\theta}_{D_{r+1}} = \sum\limits_{c}^C{\frac{m_c}{m}\varepsilon^\theta_{D_c}}$;
    \State $\varepsilon^{\theta}_{P_{r+1}} = \sum\limits_{c}^C{\frac{m_c}{m}\varepsilon^\theta_{P_c}}$;
    \item[]
\EndFor
\item[]
\Procedure{$LocalUpdate$}{$c$, $\varepsilon^\theta_{D}$,$\varepsilon^\theta_{P}$} 
    \item[]
    \For{each local epoch $i$ from 1 to $E$}
        \item[]
        \State Execute Self-Supervised Training with $\varepsilon^\theta_{D}$ and $\varepsilon^\theta_{P}$ on AV $c$;
        \State Evaluate updated $\varepsilon^\theta_{D}$ and $\varepsilon^\theta_{P}$ with Validation Dataset on AV $c$;
        \item[]
    \EndFor
    \item[]
    \State return updated $\varepsilon^\theta_{D}$ and $\varepsilon^\theta_{P}$;
    \item[]
\EndProcedure
\item[]
\Ensure Updated Global DepthNet $\varepsilon^\theta_D(I) = D$; 
\Ensure Updated Global PoseNet $\varepsilon^\theta_P(I_i,I_j) = P_{i\rightarrow j}$;
\item[]
\end{algorithmic}
\end{algorithm}

\section{Evaluation Experiments}
\label{sec:experiments}

\subsection{Datasets and Scenarios}

Our evaluation experiments are conducted with the publicly available KITTI dataset~\cite{kittydepth}, which contains monocular images and 3D scans from scenes captured by cameras and sensors mounted on top of a moving vehicle. Following the approach of~\cite{sc_depthv3, bian2021tpami, bian2019neurips}, we also adopt Eigen's split~\cite{eigen2014depth}, with the maximum depth set to 80 meters and images resized to a resolution of 832 x 256 pixels for training.

In our experiments, we assume that the 34 drives present in the training dataset correspond to distinct AVs (although some were collected by the same vehicle in different drives). Based on this assumption, we characterize three base experimental scenarios: Centralized Training, Federated Training with IID samples, and Federated Training with Non-IID samples.

\subsubsection{Centralized Training (CT)} All vehicles upload their samples to a central server that will train the depth prediction model and distribute the final version to all participants.

\subsubsection{Federated Training with IID samples (FT-IID)} The train samples are randomly distributed across the participants, preserving an equal number of samples across all participants, as depicted in Fig.~\ref{fig:samples_by_participant}. All participants share all validation samples, acting as a gold standard.
Each participant trains their local model, which is initialized with the downloaded global model, using their random subset of train samples and computes the validation losses against the gold standard at the end of every epoch. After each FL round, each participant (that was selected for that round) uploads its local model to the aggregation server, which computes the FedAvg, and then distributes the updated global model to all participants.

\subsubsection{Federated Training with Non-IID samples (FT-NIID)} 
\begin{figure}[htb]
    \centering
    \captionsetup[subfloat]{font=tiny} 
    \subfloat[]{\includegraphics[trim={0.1cm 0.4cm 2cm 2.5cm}, clip, width=5cm]{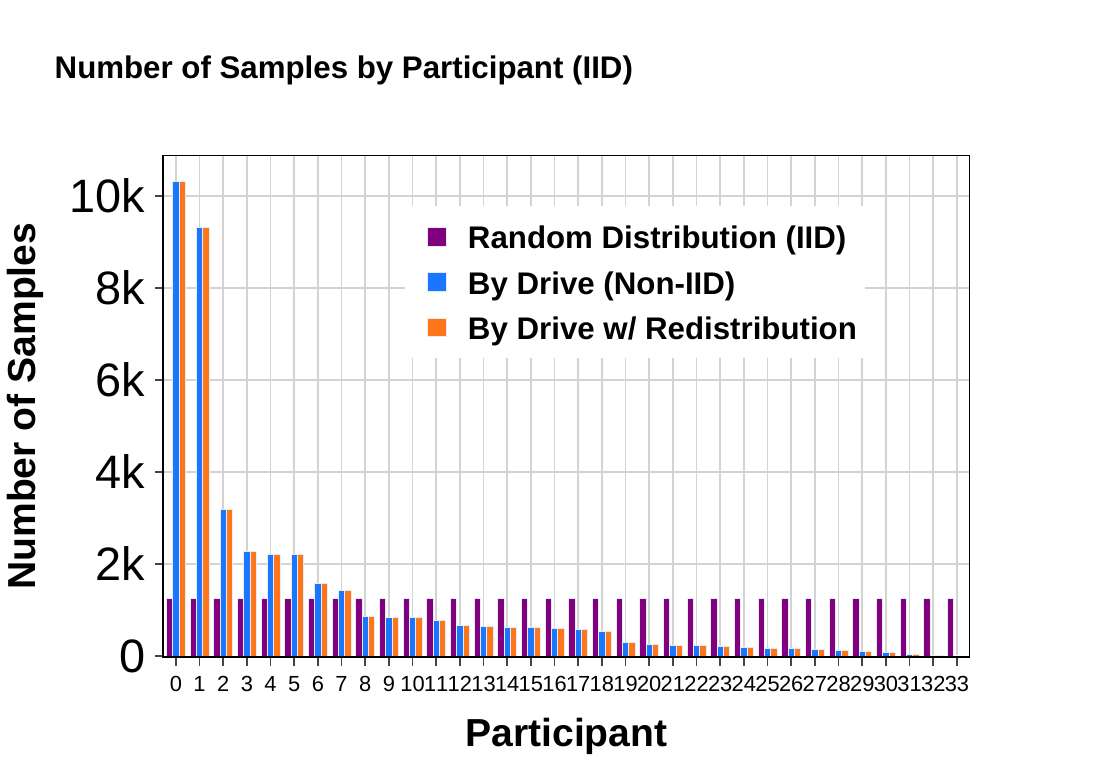}}\quad
    \subfloat[]{\includegraphics[trim={0.1cm 0.3cm 2cm 2.5cm}, clip, width=5cm]{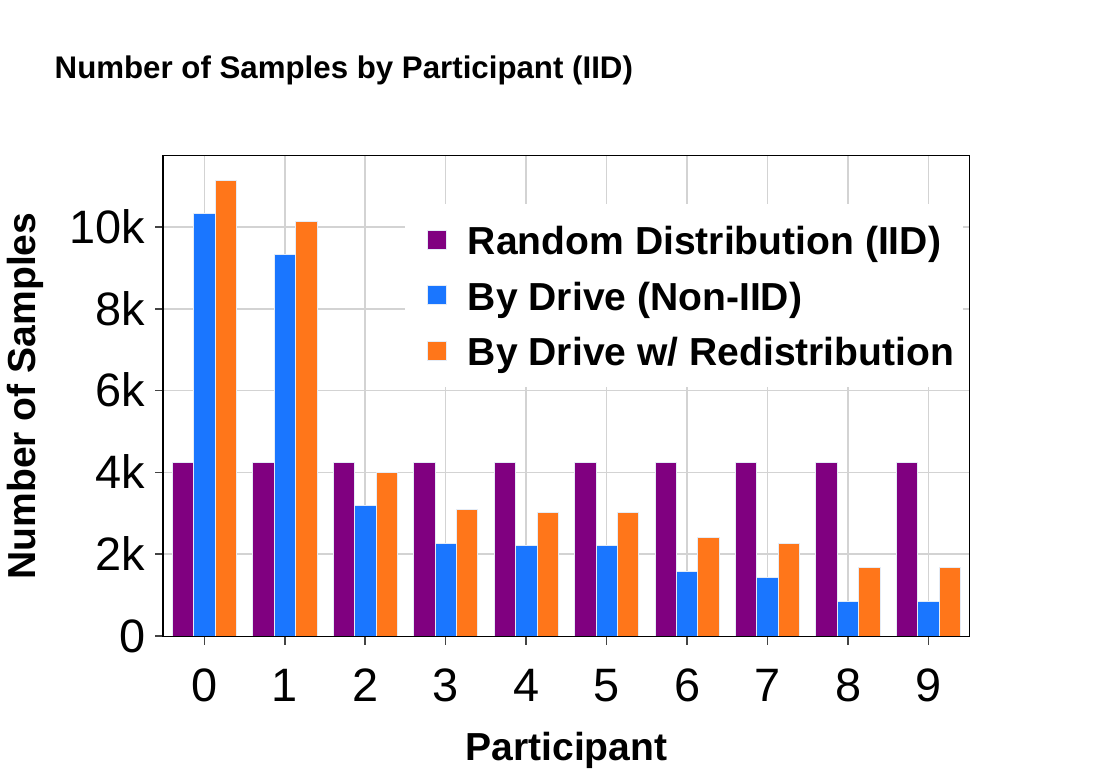}}\quad
    \subfloat[]{\includegraphics[trim={0.1cm 0.3cm 2cm 2.5cm}, clip, width=5cm]{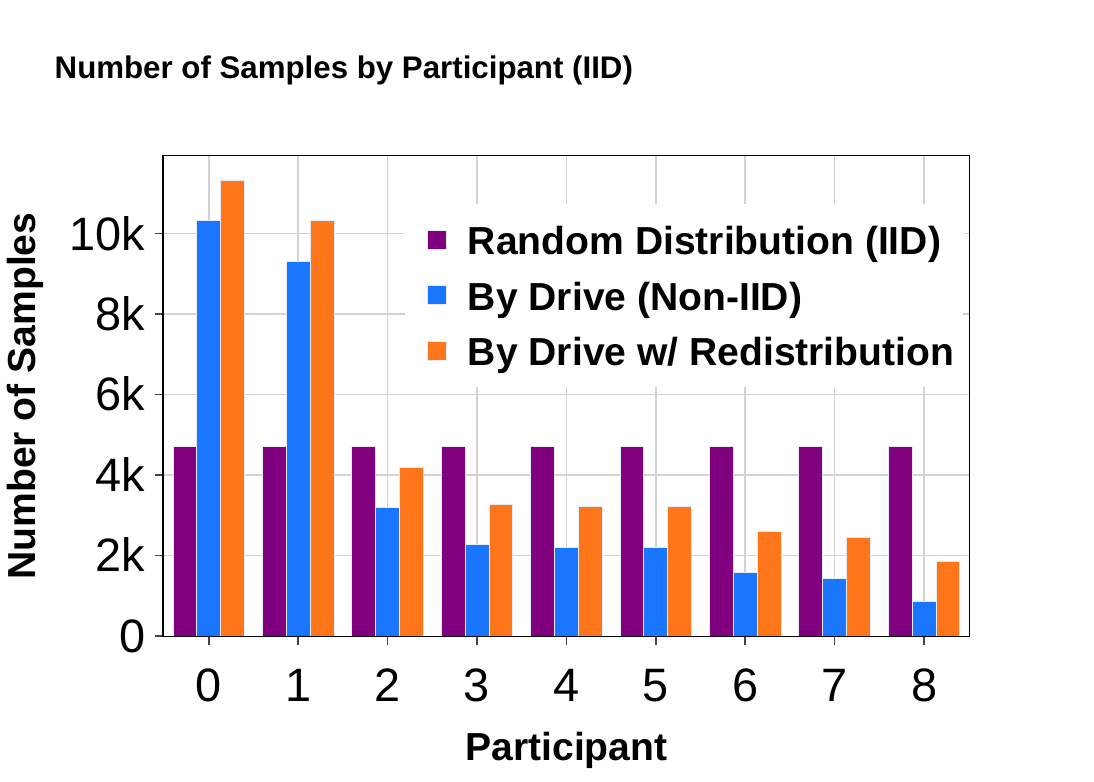}}
    \caption{Number of samples by a participant in each of the sample distribution strategies considered across 34 (a), 10 (b), and 9 (c) participants, with and without random redistribution of remaining samples.}
    \label{fig:samples_by_participant}
\end{figure}

This scenario is similar to the previous, except for the fact that the train samples are distributed according to the drives in which they were collected, reflecting the natural unbalance of the data collection. Also, since the number of samples per drive was highly skewed when selecting a subset of the 34 drives, we first picked the ones with the most train samples.
Nonetheless, the participants selected for each FL round were picked randomly, without replacement, assuring that the training would go over every participant at least once, given a sufficient number of FL rounds. In addition, to avoid creating too much advantage for the IID scenario, we randomly redistributed the remaining samples (from the participants with the least number of train samples) across the selected participants.
Thus, both the IID and the Non-IID scenarios had access to all the samples, changing only their distribution across participants. As shown in Fig.~\ref{fig:samples_by_participant}, this redistribution did not remove the great unbalance present in the original distribution by drive. However, it substantially increased the lower bound for the number of train samples.

\subsection{Metrics}
\label{sec:metrics}

For assessing the effectiveness of the final models, we adopt standard depth evaluation metrics~\cite{sc_depthv3, bian2021tpami, bian2019neurips} that include the mean absolute relative error ($AbsRel$), root mean squared error ($RMS$), root means squared log error ($RMSlog$), and accuracy under threshold ($\delta_i < 1.25^i, i = 1,2,3$), which are defined in detail in Eigens's seminal work~\cite{eigen2014depth}. Also, as in~\cite{sc_depthv3}, the predicted depth maps are multiplied by a scalar matching the median with the ground truth for evaluation.

For assessing the efficiency of the training methods, we consider the $AbsRel$ computed over the validation set during the training as our reference for effective learning, and, adapting a communication cost estimation approach proposed in~\cite{liu2021fedcpf} for FL, we formally compute its upper bound ($W_{max}$) as
\begin{equation}
W_{max} = 2T (C \times \omega^*_B).
\end{equation}
where $C$ is the total number of participants, $T$ is the total number of communication rounds (or FL rounds), and $\omega^*$ is the number of model parameters. In our formulation, we replace $\omega^*$ with $\omega^*_B$ to make it explicit that what we consider is the number of model parameters in Bytes (B). The main motivation for this is to make the comparisons with the estimated cost for CT more direct since these will be estimated based on the dataset size, which is also measured in Bytes.
Additionally, we estimate its lower bound ($W_{min}$) considering only the participants selected for training the model as,
\begin{equation}
W_{min} = 2 T (C \times F \times \omega^*_B).
\end{equation}
where $F$ is the fraction of participants that were selected for training the model locally on each FL round. In this estimate, we assume that instead of updating the global model for every participant on every round, only those that will perform local training will have access to the latest global model version.

Finally, we also analyze the number of training steps as a proxy for a computational cost estimate since these represent the number of batches the model has "seen" during the learning process (including images repeated across epochs). Formally, the number of training steps at a given epoch of the CT ($\#Steps$) can be computed as,
\begin{equation}
\#Steps = \#Epochs \times \#Batches
\end{equation}
where $\#Epochs$ is the number of epochs the model has already been trained on and $\#Batches$ is the number of batches per epoch, assuming every epoch was trained over the same number of batches. $\#Batches$ was fixed as $1k$, to enable comparison with SC-DepthV3's original results~\cite{sc_depthv3}.

For the FL scenarios, the computation of the training steps is adjusted to account for the number of participants. Thus, we define the steps in a given FT round ($\#Steps_{FL}$) as,
\begin{equation}
 \#Steps_{FL} = \sum^{\hat{T}}_{t=1} \#Steps^t_{FL},
\end{equation}
\begin{equation}
\#Steps^t_{FL} = \sum_{p \in P_t} \#Steps(p)
\end{equation}
\begin{equation}
\#Steps(p) = \#Epochs_p \times \#Batches_p
\end{equation}
where $\hat{T}$ is the number of FL rounds elapsed, $P_t$ are the participants that performed training on the round $t$ (which are not necessarily all $P$), and $\#Epochs_p$ is the number of epochs through which participant $p$ iterated over $\#Batches_p$ batches.

Thus, $\#Steps(p)$ is the number of training steps for participant $p$, and $\#Steps^t_{FL}$ is the total number of training steps for FL round $t$.
Although there might be fewer batches available for a given $p$, in the Non-IID scenarios, we have configured the training to resample the available batches randomly until the maximum number of $\#Batches_p = \#Steps$ is reached.

\subsection{Implementation Details}
\label{sec:implementationdetails}

Our FedSCDepth prototype implementation was based on SC-DepthV3~\cite{sc_depthv3} and Dec-SSL~\cite{wang2022does} source codes, which were made publicly available on GitHub by their authors. Our implementation was also shared publicly on GitHub~\footnote{\label{code}https://github.com/eltonfss/federated-sc-depth}.

As in SC-DepthV3~\cite{sc_depthv3}, the DNN implementation used PyTorch Lighting~\footnote{https://www.pytorchlightning.ai/}, with Adam optimizer, and learning rate set to $10^{-4}$. The DNN encoder was initialized using ImageNet~\cite{deng2009imagenet} pre-trained weights. As previously mentioned, the maximum number of batches per epoch was set as $1k$ and the batch size as $4$ to allow comparing CT and FT results.

FT experiments ran for 12 rounds with a total of 18 different setups resulting from the combination of the following parameter values: $C = \{10, 9\}$; $F = \{1, \frac{1}{2}, \frac{1}{3}\}$; $E = \{1, 2, 3\}$.

As in Dec-SSL~\cite{wang2022does}, the local updates were simulated on the same process in which the FedAvg aggregation was computed. Although this simulation strategy might not be sufficiently realistic for estimating all possible metrics, it does not impact the metrics we adopt for estimating the effectiveness of the trained models and the efficiency of the training.

The experiments were deployed and executed on a bare metal server with 1 x CPU i3-12100F 4,3 GHz (4 cores, 8 threads), 2 x 16GB DDR4 RAM (3200 MHz), 1 x GeForce RTX 2060 GPU with 12 GB GDDR6 RAM, and 1 x SSD M.2 2280 1TB (93GB configured as swap memory). The server was configured with Ubuntu 22.04.2 LTS, Python 3.8.15, Conda 4.12.0, Pip 22.3.1, and CUDA version 12.1.

The experiments were executed directly on the server without using any virtualization. Python dependencies were installed in a Conda environment, as described in the sources.

\subsection{Ablation Studies}

\begin{figure}[htb]
   \centering
    \captionsetup[subfloat]{font=tiny} 
    \subfloat[]{\includegraphics[trim={0.1cm 0.2cm 2cm 0.8cm}, clip, width=3.5cm]{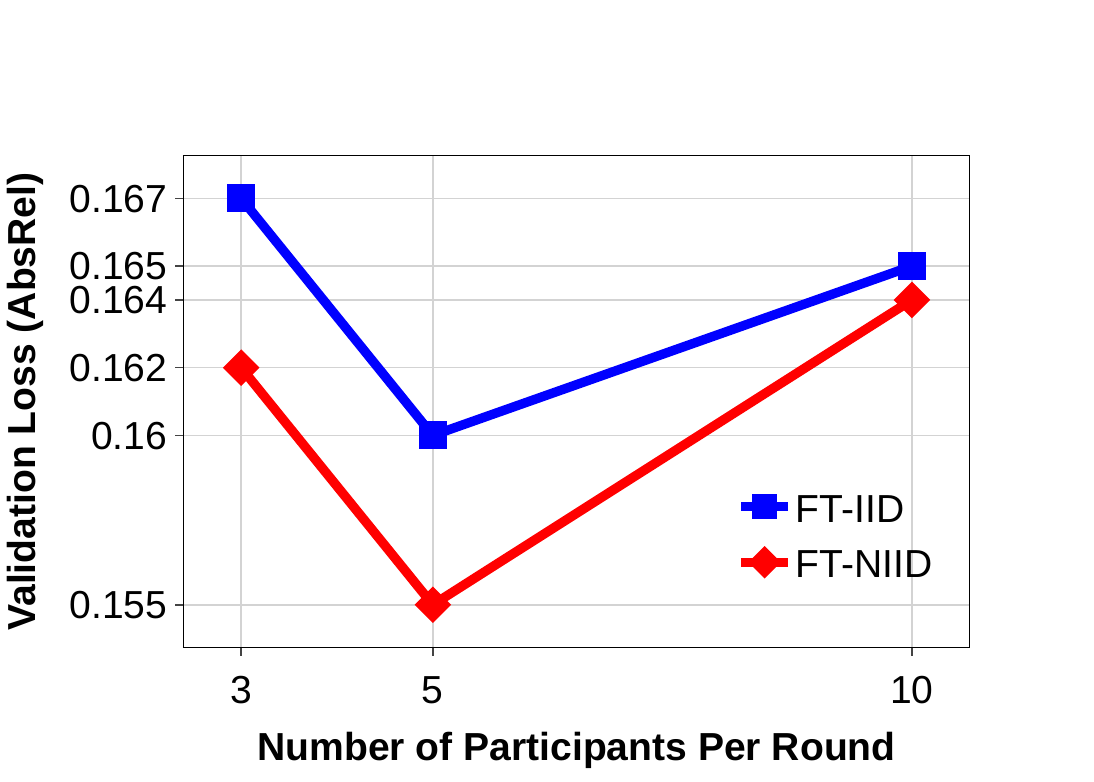}}\quad
    \subfloat[]{\includegraphics[trim={0.1cm 0.2cm 2cm 0.8cm}, clip, width=3.5cm]{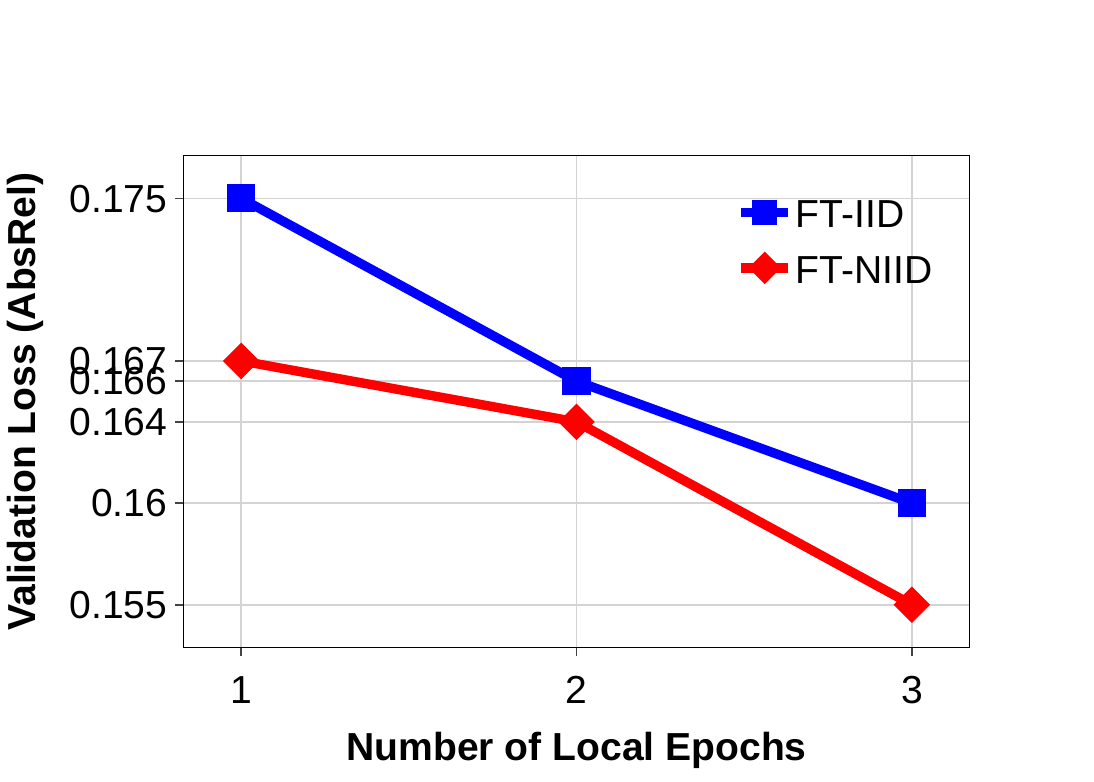}}\quad
    \subfloat[]{\includegraphics[trim={0.1cm 0.2cm 2cm 0.8cm}, clip, width=3.5cm]{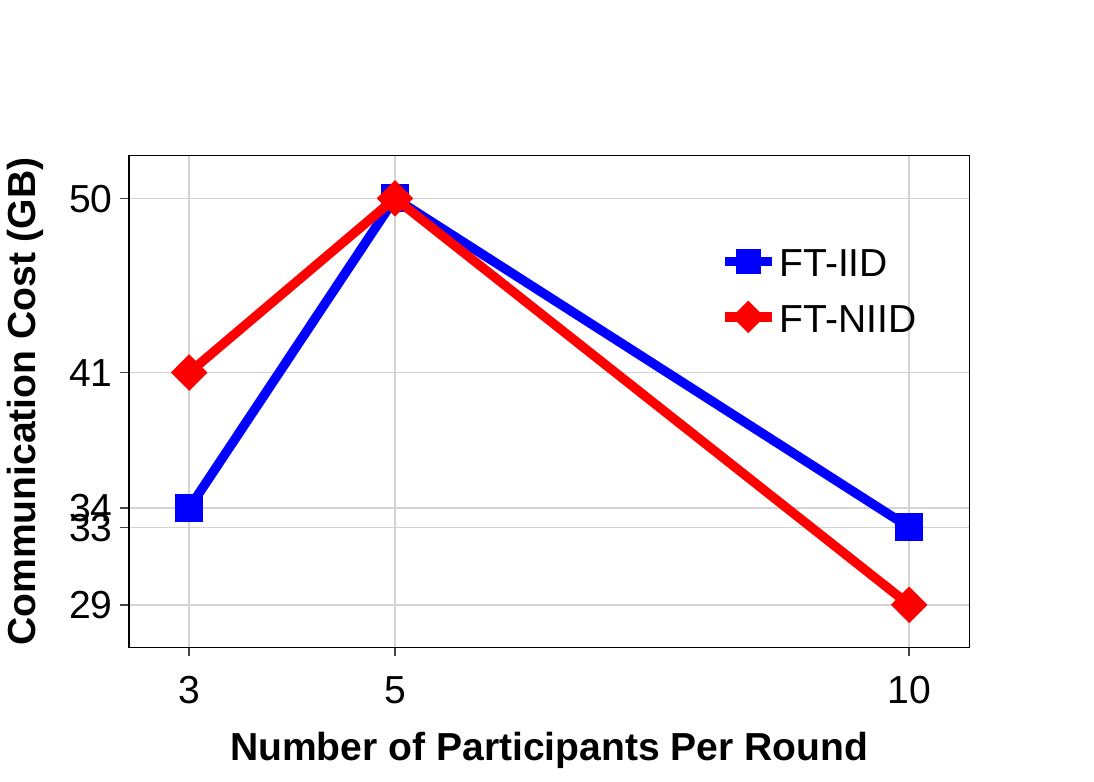}}\quad
    \subfloat[]{\includegraphics[trim={0.1cm 0.2cm 2cm 0.8cm}, clip, width=3.5cm]{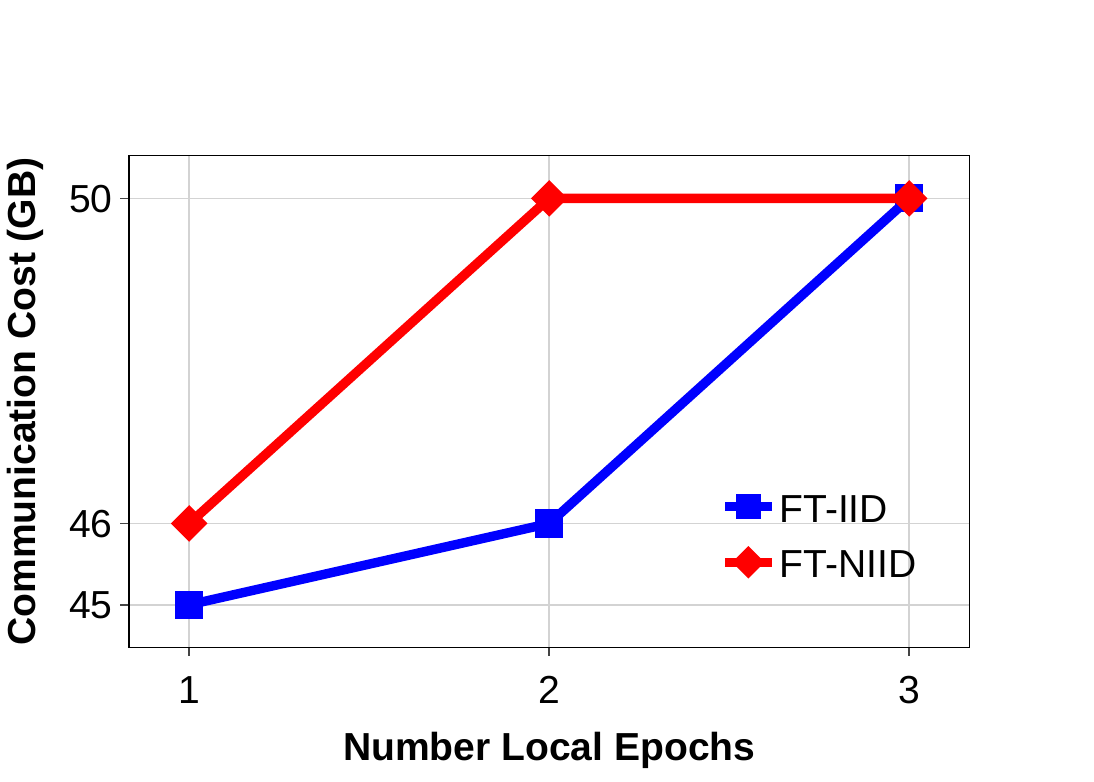}}\quad
    \subfloat[]{\includegraphics[trim={0.1cm 0.2cm 2cm 0.8cm}, clip, width=3.5cm]{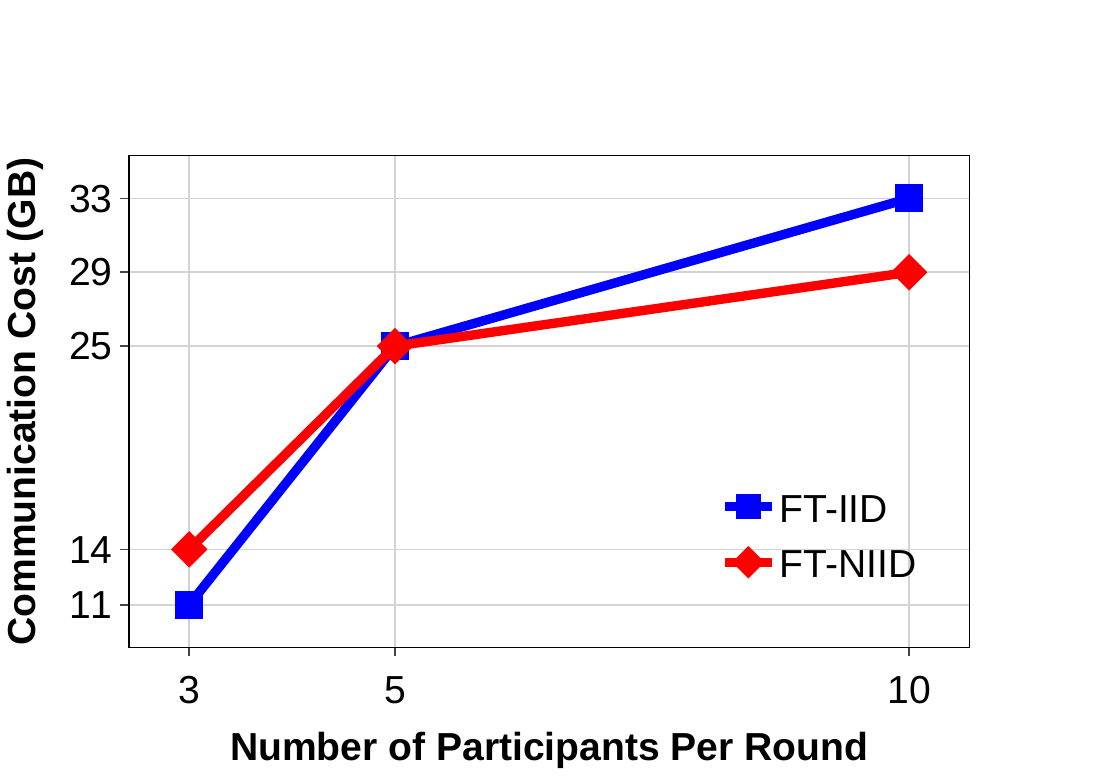}}\quad
    \subfloat[]{\includegraphics[trim={0.1cm 0.2cm 2cm 0.8cm}, clip, width=3.5cm]{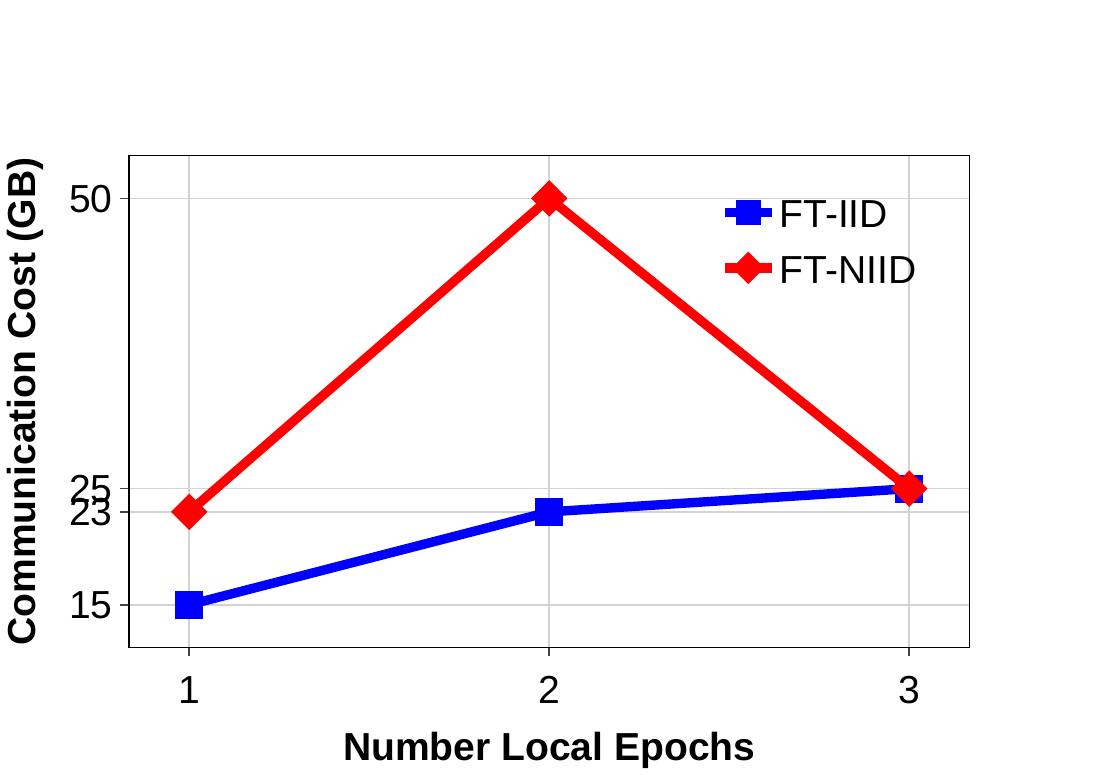}}\quad
    \subfloat[]{\includegraphics[trim={0.1cm 0.2cm 2cm 0.8cm}, clip, width=3.5cm]{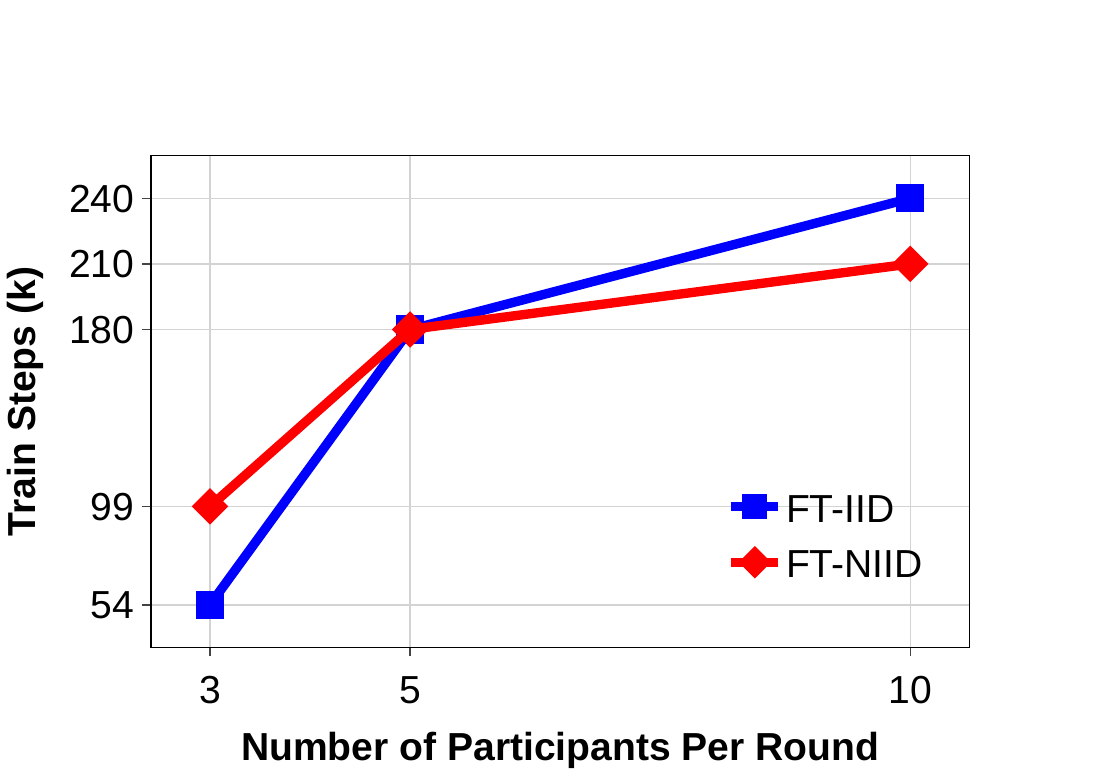}}\quad
    \subfloat[]{\includegraphics[trim={0.1cm 0.2cm 2cm 0.8cm}, clip, width=3.5cm]{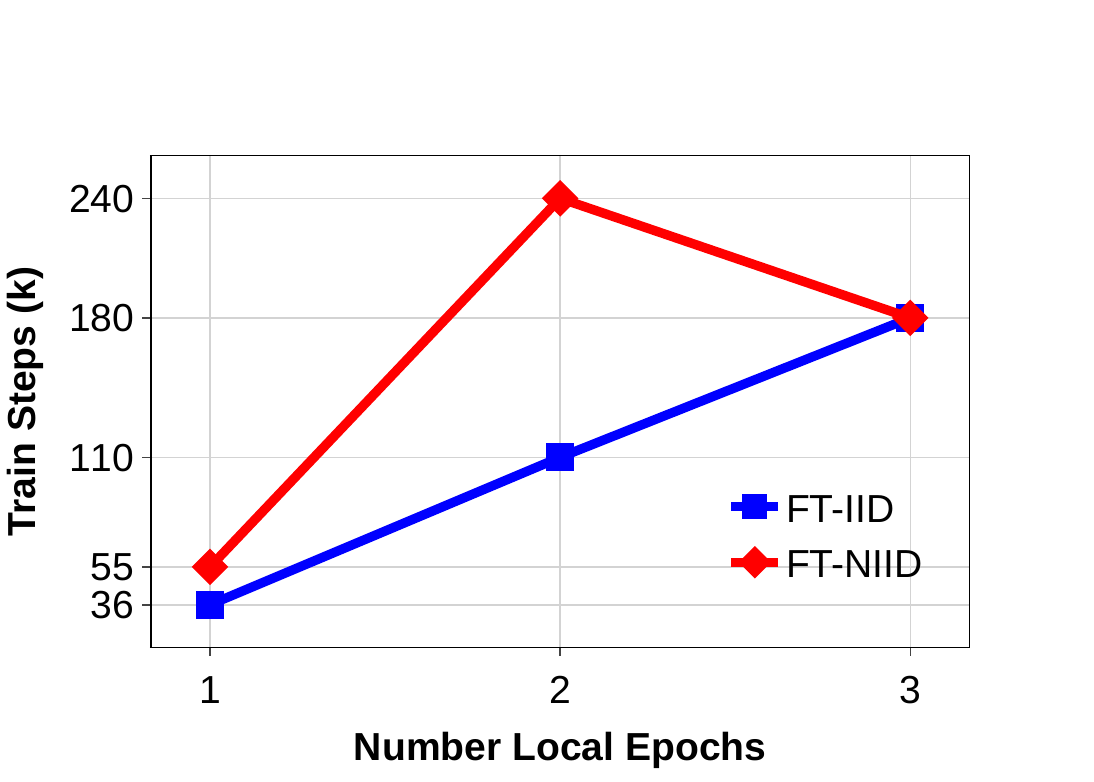}}\quad
    \caption{Lowest global validation loss (a), its estimated communication cost upper bound (c), and lower bound (e), and training steps to obtain it (g), by number of participants. Lowest global validation loss (b), its estimated communication cost upper bound (d), and lower bound (f), and training steps to obtain it (h), by number of local epochs.}
    \label{fig:metricsablation}
\end{figure}

In this section, we analyze the impact of the number of participants per round ($C*F$), the number of local training epochs ($E$), the number of communication/federation rounds ($T$), and the data heterogeneity (IID x NIID) on the lowest global validation losses ($AbsRel$) and their corresponding communication and computational costs, depicted in Figure~\ref{fig:metricsablation}.

\subsubsection{Impact of number of participants per round}
In Fig.~\ref{fig:metricsablation} (a), we find that the Validation Loss (VL) is lower when $C \times F = 5$, between $4.2\%$ to $5.5\%$ lower than the highest.
Meanwhile, in Fig.~\ref{fig:metricsablation} (c), the $W_{max}$ corresponding to the best VL is lower when $C \times F = 10$, $34\%$ to $42\%$ lower than the highest, while in Fig.~\ref{fig:metricsablation} (e), $W_{min}$ is lower when $C \times F = 3$, $51.7\%$ to $66.7\%$ lower than the highest.
Finally, in Fig.~\ref{fig:metricsablation} (g), the number of training steps corresponding to the best VL is lower when $C \times F = 3$, $52.9\%$ to $77.5\%$ lower than the highest.

\subsubsection{Impact of number of local epochs}
In Fig.~\ref{fig:metricsablation} (b), we find that the best VL is lower when $E = 3$, $7.2\%$ to $8.6\%$ lower than the highest.
Meanwhile, in Fig.~\ref{fig:metricsablation} (d), the $W_{max}$ corresponding to the best VLs is lower when $E = 1$, $8\%$ to $10\%$ lower than the highest, while in Fig.~\ref{fig:metricsablation} (f), the $W_{min}$ is also lower when $E = 1$, $40\%$ to $54\%$ lower than the highest.
Finally, in Fig.~\ref{fig:metricsablation} (h), the number of training steps corresponding to the best VLs is also lower when $E = 1$, $77.1\%$ to $80\%$ lower than the highest.

\subsubsection{Impact of number of federation rounds}
\begin{figure}[htb]
   \centering
    \captionsetup[subfloat]{font=tiny} 
    \subfloat[]{\includegraphics[trim={0.1cm 0.2cm 2cm 0.8cm}, clip, width=3.5cm]{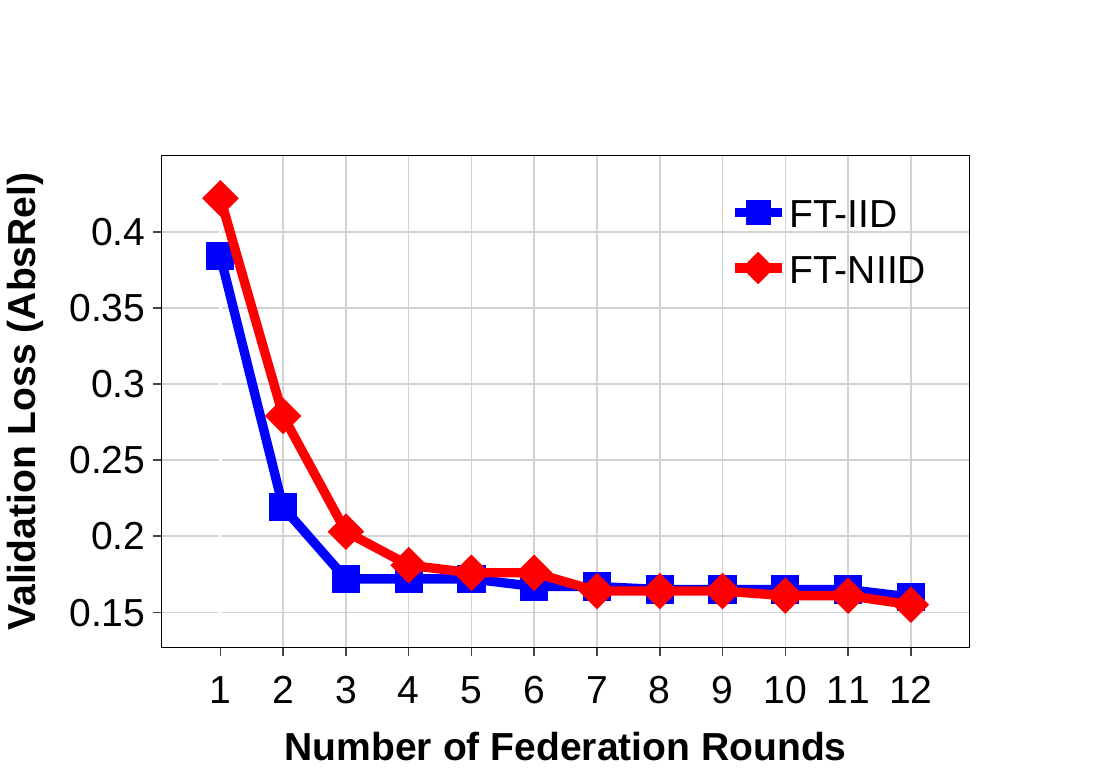}}\quad
    \subfloat[]{\includegraphics[trim={0.1cm 0.2cm 2cm 0.8cm}, clip, width=3.5cm]{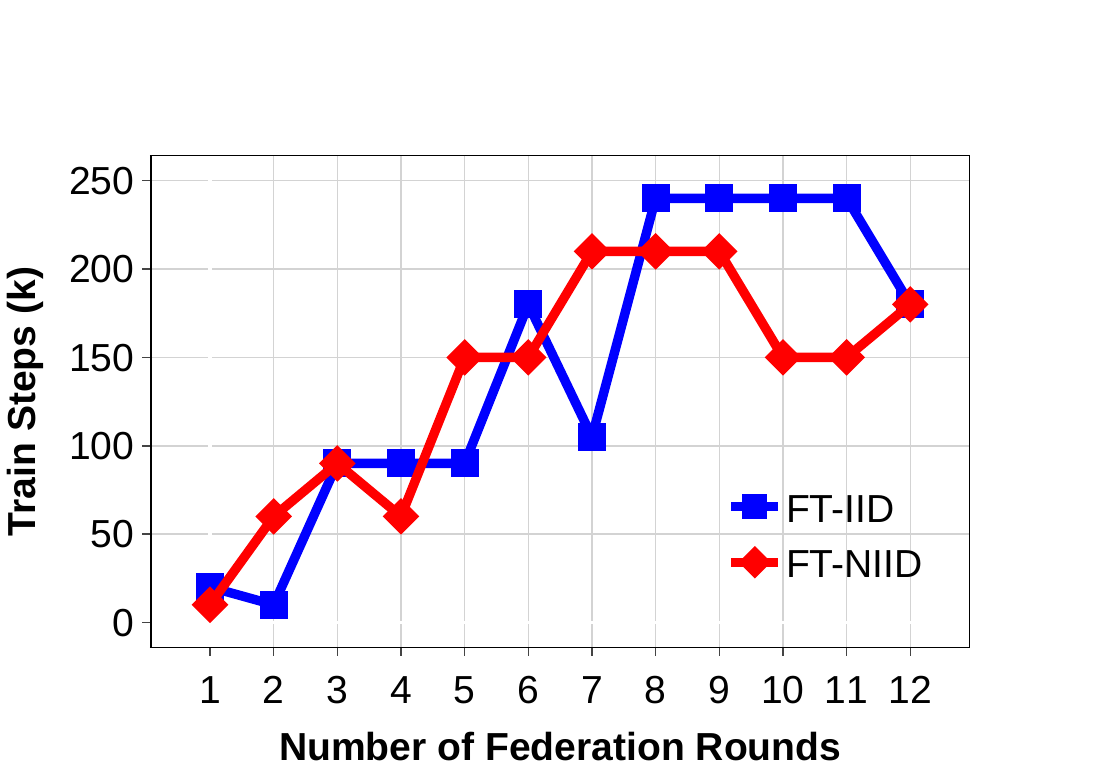}}\quad
    \subfloat[]{\includegraphics[trim={0.1cm 0.2cm 2cm 0.8cm}, clip, width=3.5cm]{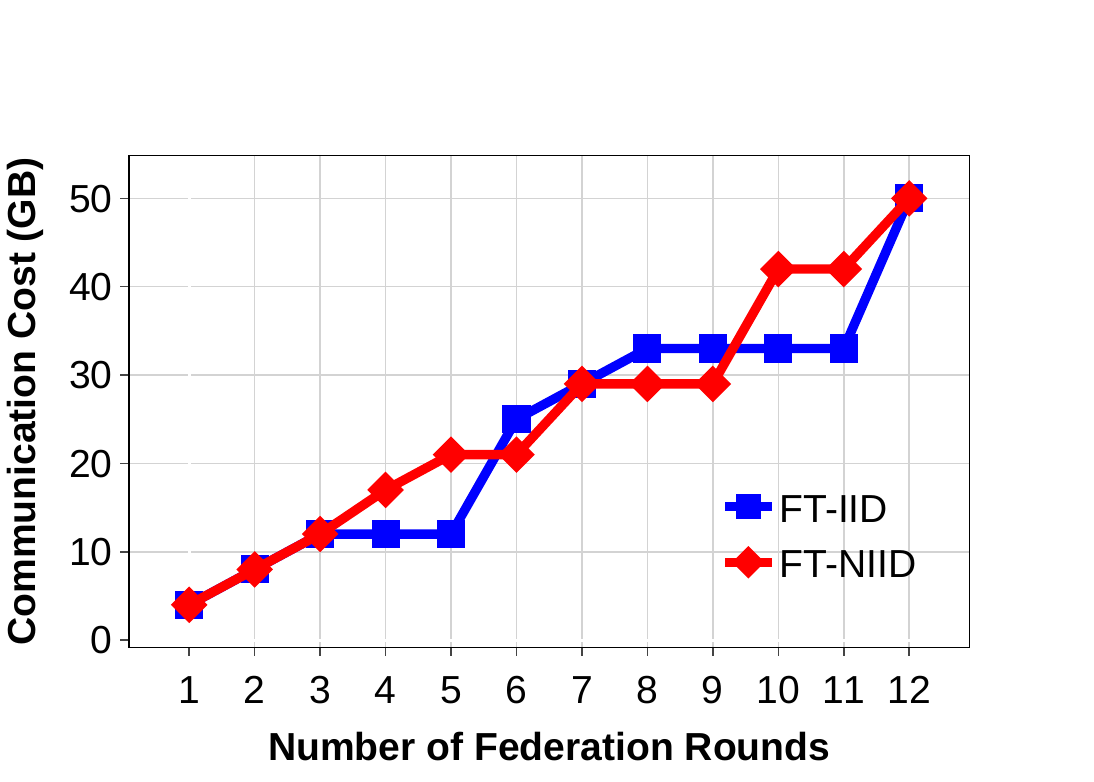}}\quad
    \subfloat[]{\includegraphics[trim={0.1cm 0.2cm 2cm 0.8cm}, clip, width=3.5cm]{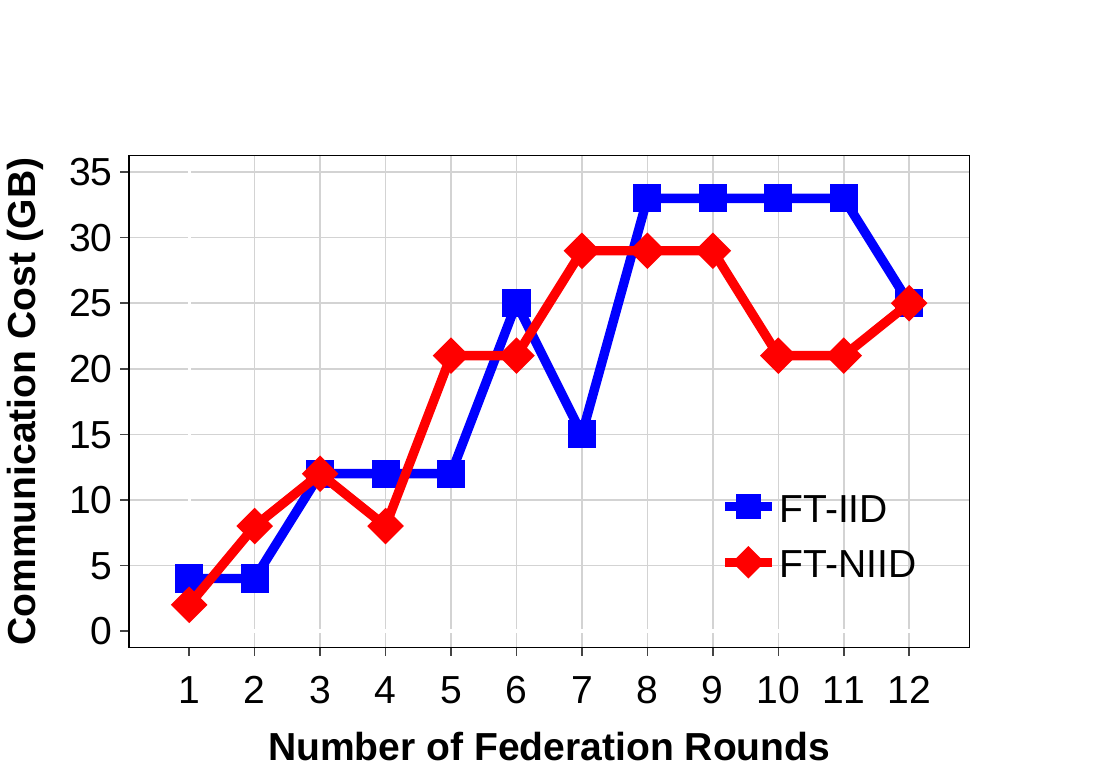}}\quad
    \caption{Lowest global validation loss (a), its communication cost upper bound (b), and lower bound (c), and steps to obtain it (d), by number of rounds.}
    \label{fig:metricsablationrounds}
\end{figure}

In Fig.~\ref{fig:metricsablationrounds} (a), we find that the best VL decreases rapidly until $T = 3$, with a modest decrease afterwards. Nonetheless, the lowest values are observed with $T = 12$. 
Meanwhile, in Fig.~\ref{fig:metricsablationrounds} (b), the number of training steps corresponding to the best VLs increases almost linearly with $T$ up to about 240k steps, at $T = 12$, while in Fig.~\ref{fig:metricsablationrounds} (c) and (d), we observe that the $W_{max}$ and $W_{min}$ corresponding to the best VLs follow a similar trend, scaling up to about 50GB and 33GB, respectively.

\subsubsection{Impact of data heterogeneity}
Analyzing Fig.~\ref{fig:metricsablation} and~\ref{fig:metricsablationrounds}, we find that the proposed method showed robustness to the data heterogeneity inherent to the data collection, obtaining the lowest VL with NIID data (about $3\%$ lower than the lowest VL with IID data).
Meanwhile, the impact on communication costs was not very high since the maximum cost (about 50GB) was the same for both.
Also, the maximum number of training steps when using NIID data was the same as IID, about 240k.

\subsection{Comparison with Centralized Training}

\begin{figure}[htb]
   \centering
    \captionsetup[subfloat]{font=tiny} 
    \subfloat[]{\includegraphics[trim={0.1cm 0.2cm 1.5cm 0.8cm}, clip, width=5cm]{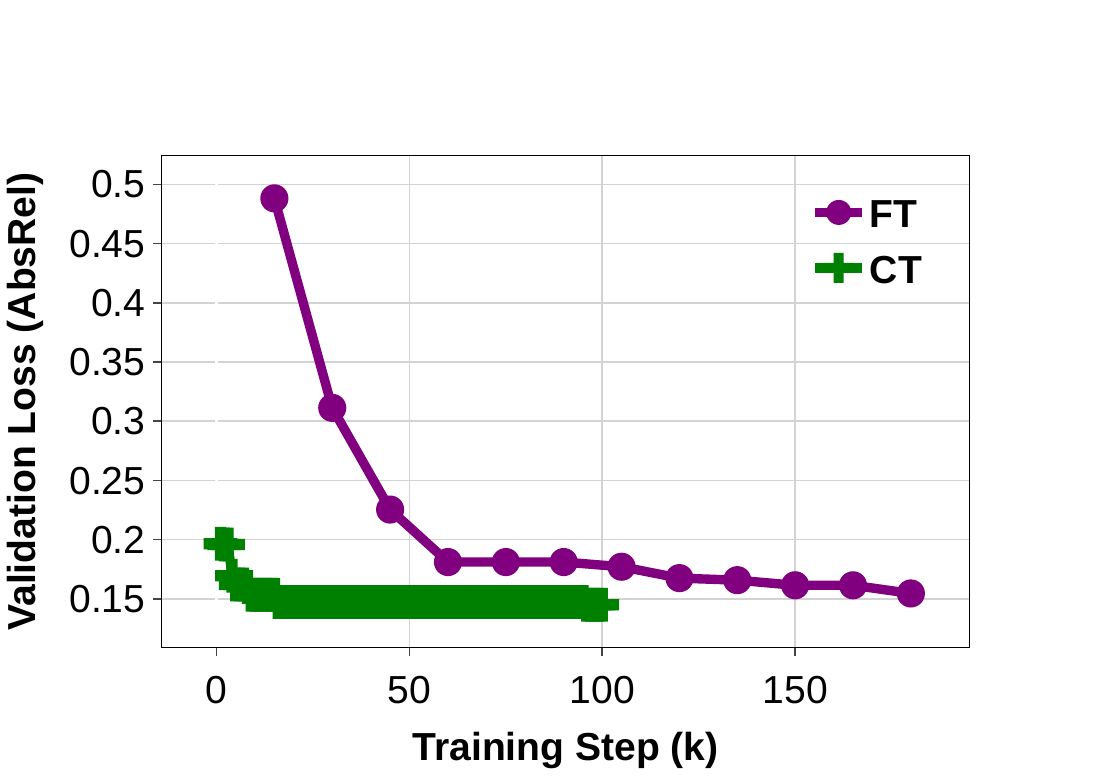}}\quad
    \subfloat[]{\includegraphics[trim={0.1cm 0.2cm 1.5cm 0.8cm}, clip, width=5cm]{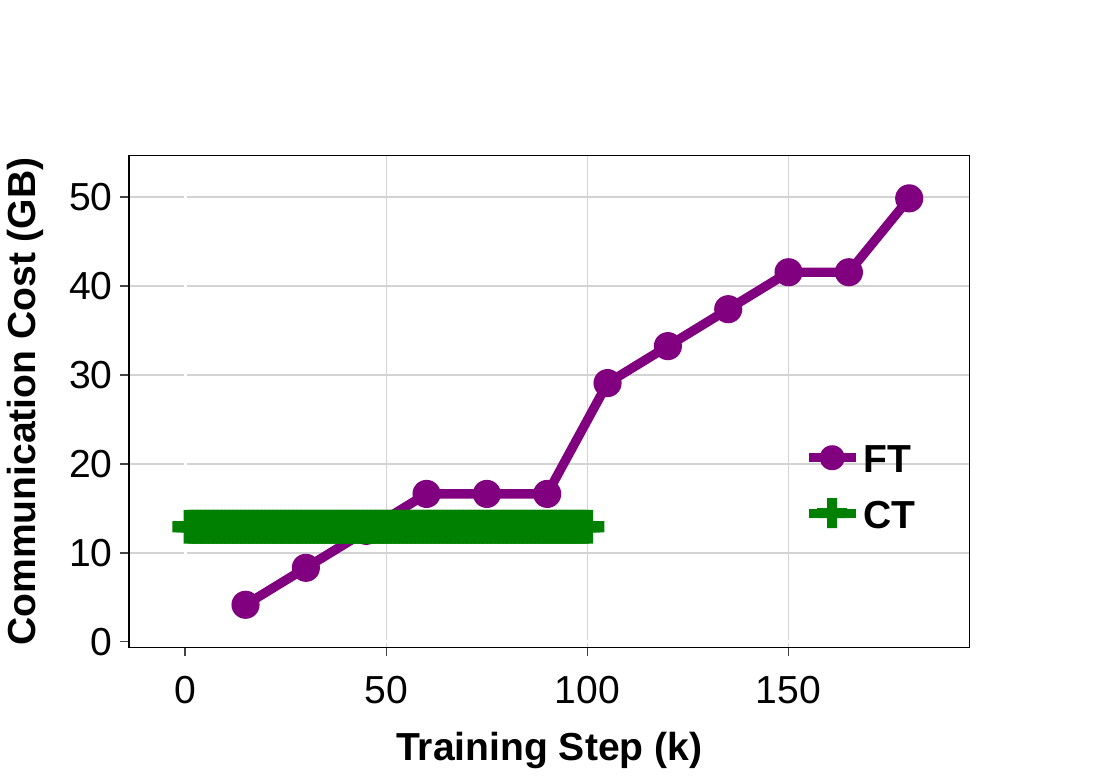}}\quad
    \subfloat[]{\includegraphics[trim={0.1cm 0.2cm 1.5cm 0.8cm}, clip, width=5cm]{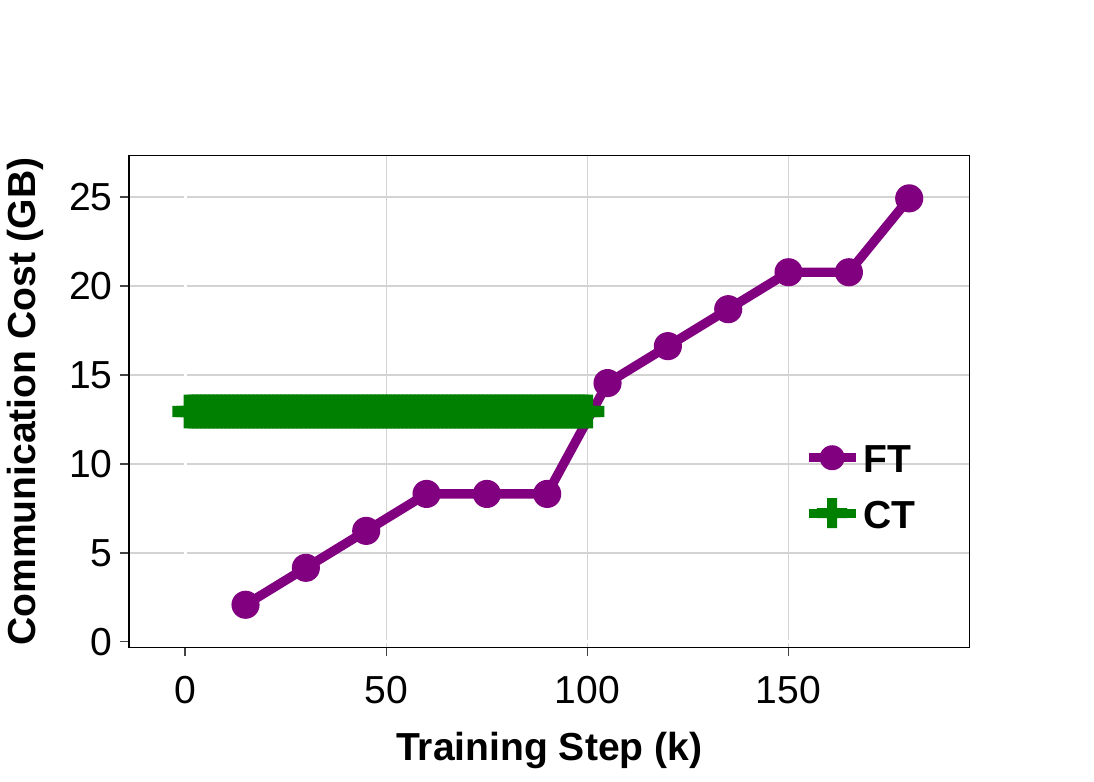}}\quad
    \caption{Lowest validation loss (a) and its estimated communication cost upper bound (b), and lower bound (c), by number of training steps. 
    }
    \label{fig:metricscomparison}
\end{figure}

After analyzing the different FT configurations, we concluded that the one that showed the best results was the one with $C \times F = 5$, $E = 3$, $T = 12$ and IID data. This was especially due to the fact that it produced the lowest VL and, although the additional communication and computation cost required for it was not negligible, it was within a reasonable value for AV use cases.
Therefore, in Fig.~\ref{fig:metricscomparison} (a), we observe that the selected FT configuration reached a VL about $7\%$ worse than the best VL obtained with CT with an additional total computational cost of about $80\%$. 
One thing to note here is that, as the VLs are computed at the end of every epoch for the CT, its first data point was obtained at 1k training steps. Meanwhile, in the FT, the first VL is computed after the first FL round is complete. Thus, the number of steps of its first data point will depend on the values of $C \times F$, $E$, and $\#Steps = 1000$.

Meanwhile, in Fig.~\ref{fig:metricscomparison} (b) and (c), we find that the final $W_{min}$ and $W_{max}$ were about $1.93\times$ and $2.85\times$ higher with FT than CT, respectively. Nonetheless, the total CT communication cost needs to be paid right at the first round, while in FT, this cost is split across 12 rounds. Considering that there would be 10 AVs involved in the data collection and training, we would have to transmit, on average, about 1.293GB per AV with CT (in the first round only) and something between 0.208GB and 0.415GB per AV with FT (on each round), which indicates that the communication cost paid by each AV on the first round would be on average $67.9\%$ to $83.9\%$ lower with FT.
Also, if we consider that in CT, the computational cost of 100k training steps has to be paid by the central server at the first round, while in FT, the computational cost of 180k training steps is shared by the 10 AVs, with an average of about 1.5k training steps being performed by each AV on each round, we conclude that FT promotes a more efficient cost distribution overall.

Finally, in Table~\ref{tbl:effectivenessmetricsregions} we observe that the efficacy metrics obtained on the test set with the best model obtained with FT were very close to the ones obtained with CT, even matching the $RMSlog$ and obtaining a slight advantage on the $SqlRel$ metrics calculated over dynamic regions (image regions classified as vehicles or pedestrians~\cite{sc_depthv3}).

\def\arraystretch{1.5}
\setlength\tabcolsep{1.5pt}
\begin{table}[htb]
	\centering
    
	\caption{Effectiveness Comparison with CT on KITTI (Eigen Split).}
	\begin{tabular}{|c|cccc|cccc|}
		\hline
		\multirow{2}{*}{\textbf{Scn.}} & \multicolumn{4}{|c|}{\textbf{Dynamic}} & \multicolumn{4}{|c|}{\textbf{Static}} \\ \cline{2-9} 
            & \textbf{$AbsRel$} & \textbf{$SqRel$} & \textbf{$RMS$} & \textbf{$RMSlog$} & \textbf{$AbsRel$} & \textbf{$SqRel$} & \textbf{$RMS$} & \textbf{$RMSlog$} \\ \hline
            FT & 0.202 & \textbf{1.933} & 7.248 & \textbf{0.282} & 0.119 & 0.723 & 4.861 & 0.177 \\ \hline
            CT & \textbf{0.191} & 2.072 & \textbf{7.111} & \textbf{0.282} & \textbf{0.106} & \textbf{0.638} & \textbf{4.275} & \textbf{0.159} \\ \hline
	\end{tabular}
	\label{tbl:effectivenessmetricsregions}
\end{table}

\subsection{Comparison with State-of-the-Art on SSL-based MDE}

In Table~\ref{tbl:metricssota}, we present the MDE efficacy metrics obtained in the test dataset with the best-performing FT configuration of the FedSCDepth. We also compare its results with the results reported by three SoTA SSL-based (centralized) MDE methods: SC-DepthV3 (SCD)~\cite{sc_depthv3}, which was the baseline of our SSL MDE component; MonoFormer (MF)~\cite{bae2023deep} and DepthFormer (DF)~\cite{guizilini2022multi}, the best performing SSL-based MDE methods in the KITTI Eigen Split~\cite{eigen2014depth}. 

\def\arraystretch{1.5}
\setlength\tabcolsep{1.5pt}
\begin{table}[htb]
    
	\centering
	\caption{Effectiveness Comparison with SoTA on KITTI (Eigen Split).}
	\begin{tabular}{|c|c|cccc|ccc|}
		\hline
            \textbf{Method} & \textbf{Resolution} & \textbf{$AbsRel$} & \textbf{$SqRel$} & \textbf{$RMS$} & \textbf{$RMSlog$} & \textbf{$\delta_1$} & \textbf{$\delta_2$} & \textbf{$\delta_3$} \\ \hline
            DF~\cite{guizilini2022multi} & $640 \times 192$ & \textbf{0.090} & \textbf{0.661} & \textbf{4.149} & \textbf{0.175} & \textbf{0.905} & \textbf{0.967} & \textbf{0.984} \\ \hline
            MF~\cite{bae2023deep} & $640 \times 192$ & 0.104 & 0.846 & 4.580 & 0.183 & 0.891 & 0.962 & 0.982 \\ \hline
            \textit{SCD~\cite{sc_depthv3}} & $832 \times 256$ & 0.118 & 0.756 & 4.709 & 0.188 & 0.864 & 0.960 & \textbf{0.984} \\ \hline
            \textit{\textbf{Ours}} & $832 \times 256$ & 0.128 & 0.803 & 5.015 & 0.197 & 0.836 & 0.956 & \textbf{0.984} \\ \hline
	\end{tabular}
	\label{tbl:metricssota}
\end{table}

Analyzing Table~\ref{tbl:metricssota}, we can observe that the $AbsRel$ obtained by FedSCDepth is about $8.5\%$, $23.1\%$, and $42.2\%$ worse than SCD, MF, and DF, respectively. Meanwhile, $\delta_3$ was the same for all except MF, and the $RMSlog$ with FedSCDepth is about $4.8\%$, $7.6\%$, and $12.6\%$ worse than with SCD, MF, and DF, respectively. Based on those results, we conclude that most of the difference between FedSCDepth and DF, which was the best performing overall, is due to the transformer-based technique employed by the latter, which produced highly superior efficacy than SCD. This difference is also visible when we compare them with MF, which presents a much closer performance to DF than SCD. Thus, our results were very close to SCD, which represents the pre-transformer SSL-based MDE SoTA and is considered our main baseline.

\section{Discussion}
\label{sec:discussion}

After analyzing the different efficiency metrics of the federated and centralized training methods, we consider the proper answer for \textit{RQ1} to be the following: FT is more efficient than CT when we accept a less strict loss threshold (such as $AbsRel$ below $0.13$). Meanwhile, to reach optimal depth prediction loss (below $0.12$), the CT will be more efficient concerning the total computational and communication costs. Nonetheless, it should be noted that while in CT, the communication cost has to be paid right at the beginning, in FT, this cost is paid across several rounds (at most, 4.15 GB of data are transferred on each round, totaling an average of 0.415 GB per participant on each round and 4.98 GB per participant after 12 rounds). 

Also, while the computational cost is entirely paid by the central server in the CT (100k training steps), this cost is shared by the participating AVs in the FT scenarios (on average, 1.5k training steps are performed by each participant at each round, totaling 18k steps of training by each participant at the 12th round).
Finally, when comparing the FT efficiency with IID and Non-IID data, it was better with IID data in most scenarios. Nonetheless, there were no significant differences in efficiency in those two FT setups overall, which is an indication that the proposed solution would perform well in realistic FT with AVs, which usually presents Non-IID data.

Meanwhile, after comparing the effectiveness of the best FT model with the SoTA, we consider the proper answer for \textit{RQ2} to be the following:
MDE models obtained with CT are more effective than those learned with FT. Nonetheless, the effectiveness lost when using FT is minimal, with the models obtained with FT reaching near SoTA performance in only 12 rounds. Also, the effectiveness obtained by the models obtained with FT is significantly better when working with NIID data for most scenarios, which indicates that this approach is highly applicable to realistic AV deployments, where data collection is typically unbalanced.

\section{Conclusion}
\label{sec:conclusion}

In this paper, we tackle the problem of monocular depth estimation for autonomous vehicles. The key to our method is using federated and self-supervised learning to collaboratively train a depth estimator using unlabeled data captured by vehicles with high effectiveness, efficiency, and privacy preservation. We evaluate a prototype implementation of this method using the KITTI dataset and show that it can achieve near-SoTA performance with a low computation cost per vehicle and a lower communication cost per round per vehicle than centralized training. Additionally, the experimental results indicate that the proposed method is robust to Non-IID data, even using simple FedAvg aggregation. Future work includes exploring other aggregation functions and optimization strategies to further reduce the proposed method's computational and communication costs, as well as evaluating its generalizability with other public benchmark datasets.

\bibliographystyle{unsrt}  
\bibliography{references}

\begin{thebibliography}{10}

\bibitem{khamis2021smart}
A~Khamis.
\newblock Smart mobility: Exploring foundational technologies and wider impacts. apress, 2021.

\bibitem{zhu2018big}
Li~Zhu, Fei~Richard Yu, Yige Wang, Bin Ning, and Tao Tang.
\newblock Big data analytics in intelligent transportation systems: A survey.
\newblock {\em IEEE Transactions on Intelligent Transportation Systems}, 2018.

\bibitem{an2011survey}
Sheng-hai An, Byung-Hyug Lee, and Dong-Ryeol Shin.
\newblock A survey of intelligent transportation systems.
\newblock In {\em IEEE CICSyN}, 2011.

\bibitem{hu2017intelligent}
Hexuan Hu, Bo~Tang, Xuejiao Gong, Wei Wei, and Huihui Wang.
\newblock Intelligent fault diagnosis of the high-speed train with big data based on deep neural networks.
\newblock {\em IEEE Transactions on Industrial Informatics}, 2017.

\bibitem{aldakkhelallah2021autonomous}
Abdulaziz Aldakkhelallah and Milan Simic.
\newblock Autonomous vehicles in intelligent transportation systems.
\newblock In {\em HCIS}, pages 185--198. Springer, 2021.

\bibitem{ming2021deep}
Yue Ming, Xuyang Meng, Chunxiao Fan, and Hui Yu.
\newblock Deep learning for monocular depth estimation: A review.
\newblock {\em Neurocomputing}, 438:14--33, 2021.

\bibitem{zhang2020monocular}
Junning Zhang, Qunxing Su, Cheng Wang, and Hongqiang Gu.
\newblock Monocular 3d vehicle detection with multi-instance depth and geometry reasoning for autonomous driving.
\newblock {\em Neurocomputing}, 403:182--192, 2020.

\bibitem{gorban2020deep}
Alexander~N Gorban, Evgeny~M Mirkes, and Ivan~Y Tyukin.
\newblock How deep should be the depth of convolutional neural networks: a backyard dog case study.
\newblock {\em Cognitive Computation}, 12(2):388--397, 2020.

\bibitem{liu2019binocular}
Fei Liu, Shubo Zhou, Yunlong Wang, Guangqi Hou, Zhenan Sun, and Tieniu Tan.
\newblock Binocular light-field: Imaging theory and occlusion-robust depth perception application.
\newblock {\em IEEE Transactions on Image Processing}, 29:1628--1640, 2019.

\bibitem{laga2019survey}
Hamid Laga.
\newblock A survey on deep learning architectures for image-based depth reconstruction.
\newblock {\em arXiv preprint arXiv:1906.06113}, 2019.

\bibitem{jing2020self}
Longlong Jing and Yingli Tian.
\newblock Self-supervised visual feature learning with deep neural networks: A survey.
\newblock {\em IEEE Transactions on Pattern Analysis and Machine Intelligence}, 2020.

\bibitem{liu2022self}
Jierui Liu, Zhiqiang Cao, Xilong Liu, Shuo Wang, and Junzhi Yu.
\newblock Self-supervised monocular depth estimation with geometric prior and pixel-level sensitivity.
\newblock {\em IEEE Transactions on Intelligent Vehicles}, 2022.

\bibitem{mcmahan2017communication}
Brendan McMahan, Eider Moore, Daniel Ramage, Seth Hampson, and Blaise~Aguera y~Arcas.
\newblock Communication-efficient learning of deep networks from decentralized data.
\newblock In {\em Artificial intelligence and statistics}. PMLR, 2017.

\bibitem{manias2021making}
Dimitrios~Michael Manias and Abdallah Shami.
\newblock Making a case for federated learning in the internet of vehicles and intelligent transportation systems.
\newblock {\em IEEE Network}, 35(3):88--94, 2021.

\bibitem{du2020federated}
Zhaoyang Du, Celimuge Wu, Tsutomu Yoshinaga, Kok-Lim~Alvin Yau, Yusheng Ji, and Jie Li.
\newblock Federated learning for vehicular internet of things: Recent advances and open issues.
\newblock {\em IEEE Open Journal of the Computer Society}, 1, 2020.

\bibitem{savazzi2021opportunities}
Stefano Savazzi, Monica Nicoli, Mehdi Bennis, Sanaz Kianoush, and Luca Barbieri.
\newblock Opportunities of federated learning in connected, cooperative, and automated industrial systems.
\newblock {\em IEEE Communications Magazine}, 59, 2021.

\bibitem{ma2022state}
Xiaodong Ma, Jia Zhu, Zhihao Lin, Shanxuan Chen, and Yangjie Qin.
\newblock A state-of-the-art survey on solving non-iid data in federated learning.
\newblock {\em Future Generation Computer Systems}, 135:244--258, 2022.

\bibitem{zhang2020federated}
Fengda Zhang, Kun Kuang, Zhaoyang You, Tao Shen, Jun Xiao, Yin Zhang, Chao Wu, Yueting Zhuang, and Xiaolin Li.
\newblock Federated unsupervised representation learning.
\newblock {\em arXiv preprint arXiv:2010.08982}, 2020.

\bibitem{zhao2021hotfed}
Juan Zhao, Ruixuan Li, Haozhao Wang, and Zijun Xu.
\newblock Hotfed: Hot start through self-supervised learning in federated learning.
\newblock In {\em 2021 IEEE HPCC/DSS/SmartCity/DependSys}, pages 149--156. IEEE, 2021.

\bibitem{zhuang2021collaborative}
Weiming Zhuang, Xin Gan, Yonggang Wen, Shuai Zhang, and Shuai Yi.
\newblock Collaborative unsupervised visual representation learning from decentralized data.
\newblock In {\em IEEE/CVF international conference on computer vision}, 2021.

\bibitem{li2022fedutn}
Simou Li, Yuxing Mao, Jian Li, Yihang Xu, Jinsen Li, Xueshuo Chen, Siyang Liu, and Xianping Zhao.
\newblock Fedutn: federated self-supervised learning with updating target network.
\newblock {\em Applied Intelligence}, pages 1--14, 2022.

\bibitem{wang2022does}
Lirui Wang, Kaiqing Zhang, Yunzhu Li, Yonglong Tian, and Russ Tedrake.
\newblock Does decentralized learning with non-iid unlabeled data benefit from self supervision?
\newblock {\em arXiv preprint arXiv:2210.10947}, 2022.

\bibitem{makhija2022federated}
Disha Makhija, Nhat Ho, and Joydeep Ghosh.
\newblock Federated self-supervised learning for heterogeneous clients.
\newblock {\em arXiv preprint arXiv:2205.12493}, 2022.

\bibitem{shi2022fedcoco}
Jiahe Shi, Yawen Wu, Dewen Zeng, Jingtong Hu, and Yiyu Shi.
\newblock Fedcoco: A memory efficient federated self-supervised framework for on-device visual representation learning.
\newblock {\em arXiv preprint arXiv:2212.01006}, 2022.

\bibitem{wu2022decentralized}
Yawen Wu, Zhepeng Wang, Dewen Zeng, Meng Li, Yiyu Shi, and Jingtong Hu.
\newblock Decentralized unsupervised learning of visual representations.
\newblock In {\em IJCAI}, 2022.

\bibitem{zhuang2022divergence}
Weiming Zhuang, Yonggang Wen, and Shuai Zhang.
\newblock Divergence-aware federated self-supervised learning.
\newblock {\em arXiv preprint arXiv:2204.04385}, 2022.

\bibitem{qu2022rethinking}
Liangqiong Qu, Yuyin Zhou, Paul~Pu Liang, Yingda Xia, Feifei Wang, Ehsan Adeli, Li~Fei-Fei, and Daniel Rubin.
\newblock Rethinking architecture design for tackling data heterogeneity in federated learning.
\newblock In {\em IEEE/CVF CVPR}, 2022.

\bibitem{mu2023fedproc}
Xutong Mu, Yulong Shen, Ke~Cheng, Xueli Geng, Jiaxuan Fu, Tao Zhang, and Zhiwei Zhang.
\newblock Fedproc: Prototypical contrastive federated learning on non-iid data.
\newblock {\em Future Generation Computer Systems}, 143:93--104, 2023.

\bibitem{park2023ms}
Sangjoon Park, Ik-Jae Lee, Jun~Won Kim, and Jong~Chul Ye.
\newblock Ms-dino: Efficient distributed training of vision transformer foundation model in medical domain through masked sampling.
\newblock {\em arXiv preprint arXiv:2301.02064}, 2023.

\bibitem{yang2023fedil}
Nan Yang, Dong Yuan, Charles~Z Liu, Yongkun Deng, and Wei Bao.
\newblock Fedil: Federated incremental learning from decentralized unlabeled data with convergence analysis.
\newblock {\em arXiv preprint arXiv:2302.11823}, 2023.

\bibitem{yang2023fedmae}
Nan Yang, Xuanyu Chen, Charles~Z Liu, Dong Yuan, Wei Bao, and Lizhen Cui.
\newblock Fedmae: Federated self-supervised learning with one-block masked auto-encoder.
\newblock {\em arXiv preprint arXiv:2303.11339}, 2023.

\bibitem{zhao2023fedusc}
Chen Zhao, Zhipeng Gao, Yang Yang, Qian Wang, Zijia Mo, and Xinlei Yu.
\newblock Fedusc: Collaborative unsupervised representation learning from decentralized data for internet of things.
\newblock {\em IEEE Internet of Things Journal}, 2023.

\bibitem{kittydepth}
Jonas Uhrig, Nick Schneider, Lukas Schneider, Uwe Franke, Thomas Brox, and Andreas Geiger.
\newblock Sparsity invariant cnns.
\newblock In {\em IEEE 3DV}, 2017.

\bibitem{jin2020practical}
Shoufeng Jin, Jiajie Yin, Mingrui Tian, Shizhe Feng, Sarkodie-Gyan Thompson, and Zhixiong Li.
\newblock Practical speed measurement for an intelligent vehicle based on double radon transform in urban traffic scenarios.
\newblock {\em MST}, 32(2), 2020.

\bibitem{eigen2014depth}
David Eigen, Christian Puhrsch, and Rob Fergus.
\newblock Depth map prediction from a single image using a multi-scale deep network.
\newblock {\em Advances in neural information processing systems}, 27, 2014.

\bibitem{bian2019neurips}
Jiawang Bian, Zhichao Li, Naiyan Wang, Huangying Zhan, Chunhua Shen, Ming-Ming Cheng, and Ian Reid.
\newblock Unsupervised scale-consistent depth and ego-motion learning from monocular video.
\newblock In {\em NeurIPS}, 2019.

\bibitem{bian2021tpami}
Jia-Wang Bian, Huangying Zhan, Naiyan Wang, Tat-Jin Chin, Chunhua Shen, and Ian Reid.
\newblock Auto-rectify network for unsupervised indoor depth estimation.
\newblock {\em IEEE Transactions on Pattern Analysis and Machine Intelligence}, 2021.

\bibitem{sc_depthv3}
Libo Sun, Jia-Wang Bian, Huangying Zhan, Wei Yin, Ian Reid, and Chunhua Shen.
\newblock Sc-depthv3: Robust self-supervised monocular depth estimation for dynamic scenes.
\newblock {\em arXiv:2211.03660}, 2022.

\bibitem{geiger2013vision}
Andreas Geiger, Philip Lenz, Christoph Stiller, and Raquel Urtasun.
\newblock Vision meets robotics: The kitti dataset.
\newblock {\em The International Journal of Robotics Research}, 32(11):1231--1237, 2013.

\bibitem{guizilini20203d}
Vitor Guizilini, Rares Ambrus, Sudeep Pillai, Allan Raventos, and Adrien Gaidon.
\newblock 3d packing for self-supervised monocular depth estimation.
\newblock In {\em IEEE/CVF CVPR}, pages 2485--2494, 2020.

\bibitem{guizilini2022multi}
Vitor Guizilini, Rareș Ambruș, Dian Chen, Sergey Zakharov, and Adrien Gaidon.
\newblock Multi-frame self-supervised depth with transformers.
\newblock In {\em CVPR}, 2022.

\bibitem{bae2023deep}
Jinwoo Bae, Sungho Moon, and Sunghoon Im.
\newblock Deep digging into the generalization of self-supervised monocular depth estimation.
\newblock In {\em AAAI}, 2023.

\bibitem{saeed2020federated}
Aaqib Saeed, Flora~D Salim, Tanir Ozcelebi, and Johan Lukkien.
\newblock Federated self-supervised learning of multisensor representations for embedded intelligence.
\newblock {\em IEEE Internet of Things Journal}, 8(2), 2020.

\bibitem{van2020towards}
Bram van Berlo, Aaqib Saeed, and Tanir Ozcelebi.
\newblock Towards federated unsupervised representation learning.
\newblock In {\em EdgeSys}, pages 31--36, 2020.

\bibitem{servetnyk2020unsupervised}
Mykola Servetnyk, Carrson~C Fung, and Zhu Han.
\newblock Unsupervised federated learning for unbalanced data.
\newblock In {\em IEEE GLOBECOM}, pages 1--6, 2020.

\bibitem{cifar}
Alex Krizhevsky.
\newblock Learning multiple layers of features from tiny images.
\newblock {\em University of Toronto}, 05 2012.

\bibitem{deng2009imagenet}
Jia Deng, Wei Dong, Richard Socher, Li-Jia Li, Kai Li, and Li~Fei-Fei.
\newblock Imagenet: A large-scale hierarchical image database.
\newblock In {\em CVPR}, 2009.

\bibitem{godard2017unsupervised}
Cl{\'e}ment Godard, Oisin Mac~Aodha, and Gabriel~J Brostow.
\newblock Unsupervised monocular depth estimation with left-right consistency.
\newblock In {\em IEEE/CVF CVPR}, 2017.

\bibitem{laina2016deeper}
Iro Laina, Christian Rupprecht, Vasileios Belagiannis, Federico Tombari, and Nassir Navab.
\newblock Deeper depth prediction with fully convolutional residual networks.
\newblock In {\em IEEE 3DV}, pages 239--248, 2016.

\bibitem{ronneberger2015u}
Olaf Ronneberger, Philipp Fischer, and Thomas Brox.
\newblock U-net: Convolutional networks for biomedical image segmentation.
\newblock In {\em MICCAI}. Springer, 2015.

\bibitem{zhou2017unsupervised}
Tinghui Zhou, Matthew Brown, Noah Snavely, and David~G Lowe.
\newblock Unsupervised learning of depth and ego-motion from video.
\newblock In {\em CVPR}, 2017.

\bibitem{he2016deep}
Kaiming He, Xiangyu Zhang, Shaoqing Ren, and Jian Sun.
\newblock Deep residual learning for image recognition.
\newblock In {\em IEEE/CVF CVPR}, pages 770--778, 2016.

\bibitem{godard2019digging}
Cl{\'e}ment Godard, Oisin Mac~Aodha, Michael Firman, and Gabriel~J Brostow.
\newblock Digging into self-supervised monocular depth estimation.
\newblock In {\em CVPR}, 2019.

\bibitem{ahuja2021deep}
Sakshi Ahuja, Bijaya~Ketan Panigrahi, Nilanjan Dey, Venkatesan Rajinikanth, and Tapan~Kumar Gandhi.
\newblock Deep transfer learning-based automated detection of covid-19 from lung ct scan slices.
\newblock {\em Applied Intelligence}, 51(1), 2021.

\bibitem{ramzan2020deep}
Farheen Ramzan, Muhammad Usman~Ghani Khan, Asim Rehmat, Sajid Iqbal, Tanzila Saba, Amjad Rehman, and Zahid Mehmood.
\newblock A deep learning approach for automated diagnosis and multi-class classification of alzheimer’s disease stages using resting-state fmri and residual neural networks.
\newblock {\em Journal of medical systems}, 44(2):1--16, 2020.

\bibitem{berga2020disentanglement}
David Berga, Marc Masana, and Joost Van~de Weijer.
\newblock Disentanglement of color and shape representations for continual learning.
\newblock {\em arXiv preprint arXiv:2007.06356}, 2020.

\bibitem{kuznietsov2021comoda}
Yevhen Kuznietsov, Marc Proesmans, and Luc Van~Gool.
\newblock Comoda: Continuous monocular depth adaptation using past experiences.
\newblock In {\em IEEE/CVF WACV}, pages 2907--2917, 2021.

\bibitem{wang2004image}
Zhou Wang, Alan~C Bovik, Hamid~R Sheikh, and Eero~P Simoncelli.
\newblock Image quality assessment: from error visibility to structural similarity.
\newblock {\em IEEE Transactions on Image Processing}, 13(4):600--612, 2004.

\bibitem{ranjan2019competitive}
Anurag Ranjan, Varun Jampani, Lukas Balles, Kihwan Kim, Deqing Sun, Jonas Wulff, and Michael~J Black.
\newblock Competitive collaboration: Joint unsupervised learning of depth, camera motion, optical flow and motion segmentation.
\newblock In {\em CVPR}, 2019.

\bibitem{liu2021fedcpf}
Su~Liu, Jiong Yu, Xiaoheng Deng, and Shaohua Wan.
\newblock Fedcpf: An efficient-communication federated learning approach for vehicular edge computing in 6g communication networks.
\newblock {\em IEEE Transactions on Intelligent Transportation Systems}, 23(2):1616--1629, 2021.

\end{thebibliography}

\end{document}